\documentclass{article}

\renewcommand{\hat}{\widehat}
\usepackage[final,nonatbib]{neurips_2023}

\usepackage{graphicx}
\usepackage{tikz}
\usepackage{bm}
\usepackage{pgfplots}
\usepackage[utf8]{inputenc} 
\usepackage[T1]{fontenc}    
\usepackage{hyperref}       
\usepackage{url}            
\usepackage{booktabs}       
\usepackage{amsfonts}       
\usepackage{nicefrac}       
\usepackage{microtype}      
\usepackage{xcolor}         
\usepackage{float}
\usepackage[ruled, vlined]{algorithm2e}
\usepackage{enumerate}
\usepackage{multirow}
\usepackage{subcaption}
\usepackage{apptools}
\usepackage[resetlabels]{multibib}
\newcites{main}{References}
\newcites{appendix}{References}
\usepackage{maths}
\newcommand{\algname}{APE }
\newcommand{\algauername}{PSI-Unif-Elim }

\title{Adaptive Algorithms for Relaxed \\ Pareto Set Identification}
\author{%
	Cyrille Kone$^1$ \\
	\texttt{cyrille.kone@inria.fr}\\
	\And
	Emilie Kaufmann$^1$ \\
	\texttt{emilie.kaufmann@univ-lille.fr}\\
	\And
 	Laura Richert$^2$ \\
	\texttt{laura.richert@u-bordeaux.fr}\\
	\\
	$^1$ Univ. Lille, Inria, CNRS, Centrale Lille, UMR 9198-CRIStAL, F-59000 Lille, France \\
	$^2$ Univ. Bordeaux, Inserm, Inria, BPH, U1219, Sistm, F-33000 Bordeaux, France
}

\newcommand{\revision}[1]{#1}

\begin{document}

\maketitle

\begin{abstract}
In this paper we revisit the fixed-confidence identification of the Pareto optimal set in a multi-objective multi-armed bandit model. As the sample complexity to identify the exact Pareto set can be very large, a relaxation allowing to output some additional near-optimal arms has been studied. In this work we also tackle alternative relaxations that allow instead to identify a relevant \emph{subset} of the Pareto set. Notably, we propose a single sampling strategy, called Adaptive Pareto Exploration, that can be used in conjunction with different stopping rules to take into account different relaxations of the Pareto Set Identification problem. We analyze the sample complexity of these different combinations, quantifying in particular the reduction in sample complexity that occurs when one seeks to identify at most $k$ Pareto optimal arms. We showcase the good practical performance of Adaptive Pareto Exploration on a real-world scenario, in which we adaptively explore several vaccination strategies against Covid-19 in order to find the optimal ones when multiple immunogenicity criteria are taken into account.
\end{abstract}

\section{Introduction}

\renewcommand{\hat}{\widehat}
\renewcommand{\tilde}{\widetilde}
In a multi-armed bandit model, an agent sequentially collects samples from several unknown distributions, called arms, in order to learn about these distributions (pure exploration), possibly under the constraint to maximize the samples collected, viewed as rewards (regret minimization). These objectives have been extensively studied for different types of univariate arms distributions \cite{lattimore_bandit_2020}. In this paper, we consider the less common setting in which arms are multi-variate distributions. We are interested in the \emph{Pareto Set Identification} (PSI) problem. In this pure exploration problem, the agent seeks to identify the arms that are \emph{(Pareto) optimal}, i.e. such that their expected values for all objectives are not uniformly worse than those of another arm. 

We formalize this as a fixed-confidence identification problem: in each round $t$ the agent selects an arm $A_t$ using an adaptive \emph{sampling rule} and observes a sample $\bX_t \in \bR^D$ from the associated distribution. It further uses an adaptive stopping rule $\tau$ to decide when to stop sampling and output a set of arms $\hat{S}_{\tau}$ which is her guess  for (an approximation of) the true Pareto set $\cS^\star$. Given a risk parameter $\delta \in (0,1)$, this guess should be correct with high probability, e.g. satisfy $\bP(\hat{S}_{\tau} = \cS^\star)\geq 1 -  \delta$ for exact Pareto set identification, while requiring a small \emph{sample complexity} $\tau$. This generalizes the well-studied fixed-confidence Best Arm Identification (BAI) problem \cite{even-dar_action_nodate,kalyanakrishnan_pac_2012,garivier_optimal_2016} to multiple objectives. 

Our motivation to study multi-objective adaptive identification stems from the design of adaptive early-phase clinical trials. In phase I/II trials, the effects of a molecule in humans are explored, and several biological endpoints may be assessed at the same time as indicative markers of efficacy. In particular, in the context of vaccine development, early-phase trials usually assess multiple immunogenicity endpoints (i.e. various markers of the effects of the vaccine on the immune system, such as different facets of antibody responses or other immune parameters). In the absence of a known correlate of protection during early clinical development, these endpoints may not have a clear \emph{a priori} hierarchy, may not all be correlated, which makes an examination of the Pareto set of different vaccinal strategies particularly relevant. In addition, given the availability of various vaccine platforms (such as mRNA vaccines, viral-vector vaccines, protein vaccines), as exemplified by Covid-19 vaccines, there may be a need to adaptively screen the various resulting vaccine strategies to select the most promising ones.
Apart from clinical trials, active Pareto Set Identification can be meaningful in many real-word contexts, and we refer the reader to the various examples given by \cite{zuluaga_active_2013}, such as in hardware or software design. \revision{Other applications include A/B/n testing for marketing or online recommender systems in which it is common to jointly optimize multiple (possibly conflicting) objectives such as user behavioral metrics (e.g. clicks, streams, dwell time, etc), supplier exposure objectives (e.g. diversity) and platform centric objectives (e.g. promotions) \cite{spotify}.}

For many applications, the sample complexity of exact PSI can be prohibitive, either when there are many close to optimal arms or when the Pareto set is very large, and different relaxations have been considered in the literature \cite{auer_pareto_2016,zuluaga_e-pal_2016}. Going back to our \revision{motivation}, in an adaptive trial that aims at pre-selecting a certain number of treatments or vaccine strategies for further investigations in clinical trials, practical constraints (the cost and feasibility of the trials) impose a constraint on the maximal number of interesting arms that can be identified. This motivates the introduction of a new setting where the agent is asked to identify \emph{at most} \(k\) Pareto optimal arms. Interestingly the sampling rule that we propose for this setting can be used to solve (some generalizations of) other relaxations considered in the literature.

\paragraph{\bfseries Related work} The work most closely related to ours is that of Auer et al. \cite{auer_pareto_2016}, who propose a relaxation, which we refer to as $\varepsilon_1$-PSI: their algorithm returns a set $\hat S$ that contains w.h.p. all the Pareto optimal arms and possibly some sub-optimal arms, which when increased by $\eps_1$ coordinate-wise become Pareto optimal. For arms that have sub-Gaussian marginals, they provide an instance-dependent sample complexity bound scaling with some notion of sub-optimality gap for each arm. The work of Zaluaga et al. \cite{zuluaga_active_2013,zuluaga_e-pal_2016} studies a structured variant of fixed-confidence PSI in which the means are regular functions of arms' descriptors. They use Gaussian process modeling and obtain worse-case sample complexity bounds. In particular \cite{zuluaga_e-pal_2016} considers the identification of an $\varepsilon$-cover of the Pareto set, which is a representative subset of the $(\varepsilon)$-Pareto set that will be related to our $(\varepsilon_1,\varepsilon_2)$-PSI criterion. The algorithms of \cite{auer_pareto_2016} and those of \cite{zuluaga_active_2013,zuluaga_e-pal_2016} in the unstructured setting\footnote{PAL relies on confidence intervals that follow from Gaussian process regression, but can also be instantiated with simpler un-structured confidence intervals as those used in our work and in Auer's} have the same flavor: they sample uniformly from a set of active arms and remove arms that have been found sub-optimal (or not representative). Auer et al.\cite{auer_pareto_2016} further adds an acceptation mechanism to stop sampling some of the arms that have been found (nearly-)optimal and are guaranteed not to dominate an arm of the active set. 
In this paper, we propose instead a more adaptive exploration strategy, which departs from such accept/reject mechanisms and is suited for different types of relaxation, including our novel $k$-relaxation. 
 

Adaptive Pareto Exploration (APE) leverages confidence intervals on the differences of arms' coordinates in order to identify a single arm to explore, in the spirit of the LUCB \cite{kalyanakrishnan_pac_2012} or UGapEc \cite{gabillon_best_2012} algorithms for Top-$m$ identification in (one-dimensional) bandit models. These algorithms have been found out to be preferable in practice to their competitors based on uniform sampling and eliminations \cite{kaufmann_information_2013}, an observation that will carry over to APE. Besides the multi-dimensional observations, we emphasize that a major challenge of the PSI problem with respect to e.g. Top $m$ identification is that the number of arms to identify is not known in advance. Moreover, when relaxations are considered, there are {multiple correct answers}. In the one-dimensional settings, finding optimal algorithms in the presence of multiple correct answers is notoriously hard as discussed by the authors of \cite{degenne_pure_2019}, and their lower-bound based approach becomes impractical in our multi-dimensional setting. Finally, we remark that the $k$-relaxation can be viewed as an {extension} of the problem of identifying any \(k\)-sized subset out of the best \(m\) arms in a standard bandit \cite{chaudhuri_pac_nodate}.


Beyond Pareto set identification, other interesting multi-objective bandit identification problems have been studied in the literature. For example \cite{drugan_designing_2013} propose an algorithm to identify {some} particular arms in the Pareto set through a scalarization technique \cite{miettinen_nonlinear_1998}. The idea is to turn the multi-objective pure-exploration problem into a single-objective one (unique optimal arm) by using a real-valued preference function which is only maximized by Pareto optimal arms (see e.g \cite{miettinen_nonlinear_1998} for some examples of these functions). 
In practice, a family of those functions can be used to identify many arms of the Pareto set but it is not always possible to identify the entire Pareto set using this technique (see e.g \cite{ehrgott_multicriteria_2005} for \emph{weighted sum} with a family of weights vectors). In a different direction, the authors of \cite{katz-samuels_feasible_2018} introduce the feasible arm identification problem, in which the goal is to identify the set of arms whose mean vectors belong to a known polyhedron $P\subset \bR^D$. 
In a follow up work \cite{katz-samuels_top_2019}, they propose a fixed-confidence algorithm for finding feasible arms that further maximize a given weighted sum of the objectives.  
In clinical trials, this could be used to find treatments maximizing efficacy (or a weighted sum of different efficacy indicators), under the constraint that the toxicity remains below a threshold. However, in the presence of multiple indicators of biological efficacy, choosing the weights may be difficult, and an examination of the Pareto set could be more suitable.
 Finally, some papers consider extensions of the Pareto optimality condition. The authors of \cite{ararat_vector_2021} tackle the identification of the set of non-dominated arms of any partial order defined by an $\bR^D$ polyhedral ordering cone (the usual Pareto dominance corresponds to using the cone defined by the positive orthant $\bR^D_+$), 
 and they provide worst-case sample complexity in the PAC setting. 
The work of \cite{audiffren_bandits_2017} studies the identification of the set of non-dominated elements in a \emph{partially ordered set} under the dueling bandit setting, in which the observations consists in pairwise comparison between arms. 
 \paragraph{\bfseries Outline and contributions} 
First, we formalize in Section~\ref{sec:problem_setting} different relaxations of the PSI problem: $\varepsilon_1$-PSI, as introduced by \cite{auer_pareto_2016}, $\varepsilon_1,\varepsilon_2$-PSI, of which a particular case was studied by \cite{zuluaga_e-pal_2016} and {$\varepsilon_1$-PSI-$k$}, a novel relaxation that takes as input an upper bound $k$ on the maximum number of $\varepsilon_1$-optimal arms that can be returned.  
Then, we introduce in Section~\ref{sec:algorithm} Adaptive Pareto Exploration, a simple, adaptive sampling rule which can simultaneously tackle all three relaxations, when coupled with an appropriate stopping rule that we define for each of them. In Section~\ref{sec:analysis}, we prove high-probability upper bounds on the sample complexity of APE under different stopping rules. For $\varepsilon_1$-PSI, our bound slightly improves upon the state-of-the-art. Our strongest result is the bound for {$\varepsilon_1$-PSI-$k$}, which leads to a new notion of sub-optimality gap, quantifying the reduction in sample complexity that is obtained. Then, Section~\ref{sec:num_sim} presents the result of a numerical study on synthetic datasets, one of them being inspired by a Covid-19 vaccine clinical trial. It showcases the good empirical performance of \algname compared to existing algorithms, and illustrates the impact of the different relaxations.

\section{Problem Setting}
\label{sec:problem_setting}
In this section, we introduce the \emph{Pareto Set Identification} (PSI) problem and its relaxations. 
Fix $K,D\in \bN^\star$. 
Let $\nu_1 , \dots, \nu_K$ be distributions over $\bR^D$ with means $\vmu_1, \dots, \vmu_K \in \bR^D.$ Let $\bA :=[K]:= \{ 1,\dots, K\}$ denote the set of arms. Let $\nu:=(\nu_1, \dots, \nu_K)$ and $\mathcal{X} := (\vmu_1,\dots,\vmu_K)$. 
We use boldfaced symbols for $\bR^D$ elements. 
Letting $\bX \in \bR^D, u \in \bR$, for any $d\in \{1, \dots, D\}$, $\tm{X}^d$ denotes the $d$-th coordinate of $\bX$ and $\bX + u:=(X^1+u, \dots, X^D+u)$. In the sequel, we will assume that $\nu_1, \dots, \nu_K$ have $1$-subgaussian marginals \footnote{A random variable $X$ is $\sigma-$subgaussian if for any $\lambda\in \bR$, $\bE(\exp(\lambda(X-\bE(X))) \leq \exp(\frac{\lambda^2\sigma^2}{2}).$}.  
\begin{definition}
Given two arms $i,j \in \bA$, $i$ is weakly (Pareto) dominated by $j$ (denoted by $\vmu_i \leq  \vmu_j$) if for any $d\in \{1, \dots, D\}$, $\mu_{i}^d \leq \mu_j^d$. The arm $i$ is (Pareto) dominated by $j$ ($\vmu_i \preceq \vmu_j$ or $i\preceq j$) if $i$ is weakly dominated by $j$ and there exists $d \in \{1, \dots, D\}$ such that $\mu_i^d < \mu_j^d$. The arm $i$ is strictly (Pareto) dominated by $j$ ($\vmu_i\prec \vmu_j$ or $i \prec j$) if for any $d \in \{1, \dots, D\}$, $\mu_i^d < \mu_j^d$. 
\end{definition}
For $\veps \in \bR_+^D$, the $\veps$-\emph{Pareto set} $\cS^\star_{\veps}(\cX)$ is the set of $\veps$-Pareto optimal arms, that is: 
\begin{equation*}
    \cS^\star_{\veps}(\cX) := \{ i \in \bA \tm{ s.t } \nexists j \in \bA: \vmu_i + \veps \prec \vmu_j \}. 
\end{equation*}
 In particular, $\cS_{\mathbf{0}}^\star(\cX)$ is called the  \emph{Pareto set} and we will simply write $\cS^\star(\cX)$ to denote $\cS^\star_{\mathbf{0}}(\cX)$. When it is clear from the context, we write $\cS^\star$ (or $\cS^\star_{\veps}$) to denote $\cS^\star(\cX)$ (or $\cS^\star_{\veps}(\cX)$).  By abuse of notation we write $\cS^\star_{\eps}$ when $\eps\in \bR^+$ to denote $\cS^\star_{\veps}$, with $\veps :=(\eps, \dots, \eps)$. 

In each round $t=1, 2, \dots$, the agent chooses an arm $A_t$ and observes an independent draw $\bX_t \sim \nu_{A_t}$ with $\bE(\bX_{A_t}) = \vmu_{A_t}$. We denote by $\bP_\nu$ the law of the stochastic process $(\bX_t)_{t\geq 1}$ and by $\bE_\nu$, the expectation under $\bP_\nu$. Let $\cF_t:= \sigma(A_1, \bX_1, \dots, A_t, \bX_t)$ the $\sigma$-algebra representing the history of the process. An algorithm for PSI consists in : i) a \emph{sampling rule} which determines which arm to sample at time $t$ based on history up to time $t-1$, ii) a \emph{stopping rule} $\tau$ which is a stopping time w.r.t the filtration $(\cF_t)_{t\geq 1}$ and iii) a \emph{recommendation rule} which is a $\cF_\tau$-measurable random set $\hat S_\tau$ representing the guess of the learner. The goal of the learner is to make a correct guess with high probability, using as few samples $\tau$ as possible. Before formalizing this, we introduce the different notion of correctness considered in this work, depending on parameters $\varepsilon_1\geq, \varepsilon_2 \geq 0$ and $k \in [K]$. Our first criterion is the one considered by \cite{auer_pareto_2016}.

\begin{definition}
    $\hat S \subset \bA$ is correct for $\eps_1$-PSI if $\cS^\star \subset \hat S \subset \cS^\star_{\eps_1}$. 
\end{definition}

To introduce our second criterion, we need the following definition.
\begin{definition}
    \label{def:eps-cover}
    Let $\eps_1, \eps_2 \geq 0$. A subset $S\subset \bA$ is an $(\eps_1, \eps_2)$-cover of the \emph{Pareto set} if : $S\subset \cS^\star_{\eps_1}$ and for any $i\notin S$ either $i\notin \cS^\star$  or $\exists j \in S$ such that $\vmu_i \prec \vmu_j + {\eps_2}$.
\end{definition}
The $\eps$-accurate set of \cite{zuluaga_e-pal_2016} is a  particular case of $(\eps_1, \eps_2)$-cover for which $\eps_1 = \eps_2 = \eps$. Allowing $\eps_1\neq \eps_2$ generalizes the notion of $\eps$-correct set and can be useful, e.g., in scenarios when we want to identify the exact {Pareto set} (setting $\eps_1=0$) but allow some optimal arms to be discarded if they are too close (parameterized by $\eps_2$) to another optimal arm already returned. We note however that the \emph{sparse cover} of \cite{auer_pareto_2016} is an $(\varepsilon,\varepsilon)$-cover with and additional condition that the arms in the returned set should not be too close to each over. Identifying a sparse cover from samples requires in particular to identify $\cS_{\varepsilon_1}^\star$ hence it can not be seen as a relaxation of $\varepsilon_1$-PSI.  

\begin{definition}
   $\hat S \subset \bA$ is correct for $(\eps_1, \eps_2)$-PSI if it is an $(\eps_1, \eps_2)$-cover of the Pareto set. 
\end{definition}

\begin{definition}
    $\hat S \subset \bA$ is correct for $\eps_1$-{PSI}-$k$ if either \emph{i)} $\lvert \hat S\lvert = k$ and $\hat S \subset \cS^\star_{\eps_1}$    or \emph{ii)}  $ \lvert \hat S\lvert <k$  and $\cS^\star \subset \hat S \subset \cS^\star_{\eps_1}$ holds. 
\end{definition}
Given a specified objective ($\eps_1$-PSI, $(\eps_1, \eps_2)$-PSI or $\eps_1$-PSI-$k$), and a target risk parameter $\delta \in (0,1)$, the goal of the agent is to build a $\delta$-correct algorithm, that is to guarantee that with probability larger than $1-\delta$, her guess $\hat{S}_{\tau}$ is correct for the given objective, while minimizing the number of samples $\tau$ needed to make the guess, called the \emph{sample complexity}.

We now introduce two important quantities to characterize the (Pareto) optimality or sub-optimality of the arms. For any two arms $i, j$, we let
\begin{eqnarray*}
  \m(i,j)&:=& \min_{1\leq d\leq D}(\mu_j^d - \mu_i^d), \text{ and } \M(i,j) := \max_{1\leq d\leq D} (\mu_i^d - \mu_j^d),
\end{eqnarray*}
which have the following interpretation. If $i\preceq j$, $\m(i,j)$ is the minimal quantity $\alpha \geq0$ that should be added component-wise to $\vmu_i$ so that $ \vmu_i + \boldsymbol{\alpha} \nprec \vmu_j$, $\boldsymbol{\alpha}:=(\alpha, \dots, \alpha)$.  Moreover, $\m(i,j) > 0$ if and only if $i  \prec j$. Then, for any arms $i,j$, if $i\nprec j$, $\M(i,j)$ is the minimum quantity $\alpha'$  such $\vmu_i \leq \vmu_j + \boldsymbol{\alpha'} $, $\boldsymbol{\alpha'}:=(\alpha' ,\dots, \alpha')$. We remark that $\M(i,j)<0$ if and only if $i\prec j$. Our algorithms, presented in the next section, rely on confidence intervals on these quantities.

\section{Adaptive Pareto Exploration}\label{sec:algorithm}

We describe in this section our sampling rule, Adaptive Pareto Exploration, and present three stopping and recommendation rules to which it can be combined to solve each of the proposed relaxation.
Let $T_k(t):=\sum_{s=1}^{t-1}\ind(A_s=k)$ be the number of times arm $k$ has been pulled up to round $t$ and $\vmuh_k(t):= T_k(t)^{-1} \sum_{s=1}^{T_k(t)} \bX_{k, s}$  the empirical mean of this arm at time $t$, where ${\bX}_{k, s}$ denotes the $s$-th observation drawn \iid from $\nu_k$. For any arms $i,j\in \bA$, we let \[\m(i,j,t) := \min_d (\muh_j^d(t) - \muh_i^d(t)) \ \ \text{ and } \ \ \M(i,j,t):= \max_d (\muh^d_i(t) - \mu^d_j(t)).\] The empirical Pareto set is defined as  
\begin{eqnarray*}S(t) &:=& \{i \in \bA: \nexists j \in \bA : \vmuh_i(t) \prec \vmuh_j(t) \},\\
 & = &\{i \in \bA: \forall j  \in \bA\backslash \{i\}, \Mh(i,j,t) > 0 \}\;.
\end{eqnarray*}

\subsection{Generic algorithm(s)}
Adaptive Pareto Exploration relies on a {\it lower/upper confidence bound} approach, similar to  single-objective BAI algorithms like UGapEc \cite{gabillon_best_2012}, LUCB\cite{kalyanakrishnan_pac_2012} and LUCB++ \cite{simchowitz_simulator_2017}. 
These three algorithms identify in each round two contentious arms: $b_t$: a current guess for the optimal arm (defined as the empirical best arm or the arm with the smallest upper bound on its sub-optimality gap), $c_t$: a contender of this arm; the arm which is the most likely to outperform $b_t$ (in all three algorithms, it is the arm with the largest upper confidence bound in $[K]\backslash \{b_t\}$). Then, either both arms are pulled (LUCB, LUCB++) or the least explored among $b_t$ and $c_t$ is pulled (UGapEc). The originality of our sampling rule lies in how to appropriately define $b_t$ and $c_t$ for the multi-objective setting. To define those, we suppose that there exists confidence intervals $[L_{i,j}^{d}(t,\delta),U_{i,j}^{d}(t,\delta)]$ on the difference of expected values for each pair of arms $(i,j)$ and each objective $d \in D$, such that introducing 
\begin{eqnarray}
    \cE_t := \bigcap_{i=1}^K \bigcap_{j\neq i}\bigcap_{d=1}^D   \left\{L^d_{i,j}(t, \delta) \leq \mu_i^d - \mu_j^d \leq U^d_{i,j}(t, \delta)\right\} \ \text{ and } \ \cE = \bigcap_{t =1}^{\infty} \cE_{t},\label{eq:def_good_event}
\end{eqnarray} we have $\bP(\cE)\geq 1-\delta$. Concrete choices of these confidence intervals will be discussed in Section~\ref{subsec:concrete}. 

To ease the notation, we drop the dependency in $\delta$ in the confidence intervals and further define 
\begin{eqnarray}
    \Mh^-(i,j,t) &:=& \max_d L_{i,j}^d(t) \ \ \ \ \ \text{ and } \ \ \Mh^+(i,j,t) := \max_d U^d_{i,j}(t)\\
\mh^-(i,j,t) &:=& -\Mh^+(i,j,t) \ \ \ \text{ and} \ \ \ \mh^+(i,j, t) := - \Mh^-(i,j,t).
\end{eqnarray}
\begin{restatable}{lemma}{ineqGene}\label{lem:ineq_gene} For any round $t\geq 1$, if $\cE_t$ holds, then for any $i,j \in \bA$, $\Mh^{-}(i,j,t) \leq \M(i,j) \leq \Mh^{+}(i,j,t)$ and $\mh^{-}(i,j,t) \leq \m(i,j) \leq \mh^{+}(i,j,t)$.
\end{restatable}

%
%

Noting that $\cS^\star_{\varepsilon_1} = \{ i \in \bA : \forall j \neq i, \M(i,j)+\eps_1 > 0\}$, we define the following set of arms that are likely to be $\eps_1$-Pareto optimal: 
\[\OPT^{\eps_1}(t) := \{i \in \bA: \forall  j  \in \bA\backslash \{i\},  \Mh^-(i,j,t) +\eps_1>0 \}.\]

\paragraph{Sampling rule} In round $t$, Adaptive Pareto Exploration samples $a_t$, the least pulled arm among \revision{two candidate arms} $b_t$ and $c_t$ 
given by 
\begin{eqnarray*}
  b_t &:=& \argmax_{i\in \bA \setminus \OPT^{\eps_1}(t)} \min_{j \neq i}\Mh^+(i,j,t),  \\
  c_t &:=& \argmin_{j \neq  b_t} \Mh^-(b_t, j, t)
\end{eqnarray*}


 The intuition for their definition is the following. Letting $i$ be a fixed arm, note that $\M(i,j)>0$ for some $j$, if and only if there exists a component $d$ such that $\mu_i^d > \mu_j^d$ i.e $i$ is not dominated by $j$. Moreover, the larger $\M(i,j)$, the more $i$ is non-dominated by $j$ in the sense that there exists $d$ such that $\mu_i^d\gg\mu_j^d$. Therefore, $i$ is strictly optimal if and only if  for all $j\neq i$, $\M(i,j)>0$ i.e $\alpha_i:= \min_{j\neq i}\M(i,j)>0$. And the larger $\alpha_i$, the more $i$ looks optimal in the sense that for each arm $j\neq i$, there exists a component $d_j$ for which $i$ is way better than $j$. As the $\alpha_i$ are unknown, we define $b_t$ as the maximizer of an optimistic estimate of the $\alpha_i$'s. We further restrict the maximization to arms for which we are not already convinced that they are optimal (by \autoref{lem:ineq_gene}, the arms in $\OPT^{\eps_1}(t)$ are (nearly) Pareto optimal on the event $\cE$). Then, we note that for a fixed arm $i$, $\M(i,j) < 0$ if and only if $i$ is strictly dominated by $j$. And the smaller $\M(i,j)$, \revision{the more $j$ is close to dominate $i$ (or largely dominates it): for any component $d$, $\mu_i^d - \mu_j^d$ is small (or negative)}. Thus, for a fixed arm $i$, $\argmin_{j\neq i} \M(i,j)$ can be seen as the arm which is the closest to dominate $i$ (or which dominates it by the largest margin). 
 By minimizing a lower confidence bound on the unknown quantity $\M(b_t,j)$, our contender $c_t$ can be interpreted as the arm which is the most likely to be (close to) dominating $b_t$. Gathering information on both $b_t$ and $c_t$ can be useful to check whether $b_t$ can indeed be optimal. 
 
 \revision{Interestingly, we show in Appendix~\ref{sec:bai} that for $D=1$, our sampling rule is close but not identical to the sampling rules used by existing confidence-based best arm identification algorithms.}


\paragraph{Stopping  and recommendation rule(s)} Depending on the objective, Adaptive Pareto Exploration can be plugged in with different stopping rules, that are summarized in Table~\ref{tab:stop-rec-rules} with their associated recommendations. To define those, we define for all $i \in\bA$, $\varepsilon_1,\varepsilon_2\geq 0$,   
\begin{eqnarray*}
   g_i^{\eps_2}(t):= \max_{j \neq i} \m^-(i,j,t) + \eps_2 \ind\{j \in \OPT^{\eps_1}(t)\}\quad \tm{and} \quad
   h_i^{\eps_1}(t) := \min_{j\neq i}\Mh^-(i,j,t) + {\eps_1}.  
\end{eqnarray*}
and let $g_i(t):= g^{0}_i(t)$. Introducing 
\begin{eqnarray*}
    Z_1^{\eps_1}(t) := \min_{i \in S(t)} h^{\eps_1}_i(t), \ \text{ and } \   Z_2^{\eps_1}(t):= \min_{i \in S(t)^{\complement}}  \max(g_i(t), h^{\eps_1}_i(t)),
\end{eqnarray*}
for $\varepsilon_1$-PSI, our stopping rule is $\tau_{\varepsilon_1} := \inf \{t \geq K : Z_1^{\eps_1}(t) > 0 \wedge Z_2^{\eps_1}(t) > 0 \}$ and the associated recommendation is $\cO({\tau_{\varepsilon_1}})$ where 
\begin{equation*}
    \label{eq:recrule_epsPSI}
    \cO({t}) := 
    S(t) \cup \{ i \in S(t)^\complement: \nexists j\neq i: \mh^-(i,j,t)> 0\}
\end{equation*}
consists of the current empirical {Pareto} set plus some additional arms that have not yet been formally identified as sub-optimal. Those arms should be $(\eps_1)$-Pareto optimal.

For $(\varepsilon_1,\varepsilon_2)$-PSI we define a similar stopping rule $\tau_{\eps_1,\eps_2}$ where the stopping statistics are respectively replaced with 
\begin{eqnarray*}
    Z_1^{\eps_1, \eps_2}(t) := \min_{i \in S(t)} \max(g_i^{\eps_2}(t), h^{\eps_1}_i(t)) \ \text{ and} \ Z_2^{\eps_1, \eps_2}(t):= \min_{i \in S(t)^{\complement}}  \max(g_i^{\eps_2}(t), h^{\eps_1}_i(t))
\end{eqnarray*}
with the convention $\min_{\emptyset} = +\infty$, and the recommendation is $\text{OPT}^{\varepsilon_1}(\tau_{\varepsilon_1,\varepsilon_2})$.

To tackle the $\eps_1$-PSI-$k$ relaxation, we propose to couple $\tau_{\eps_1}$ with an additional stopping condition checking whether 
$\OPT^{\eps_1}(t)$ already contains $k$ arms. That is, we stop at $\tau_{\varepsilon_1}^{k} := \min\left(\tau_{\eps_1},\tau^{k}\right)$ where 
$\tau^{k} := \inf\{t \geq K : |\OPT^{\eps_1}(t)| \geq k \}$ with associated recommendation $\OPT^{\eps_1}(\tau^{k})$. Depending of the reason for stopping ($\tau_{\eps_1}$ or $\tau^{k}$), we follow the corresponding recommendation.


\begin{table}[t]
    \centering
    \begin{tabular}{|c|c|c|c|}
    \hline 
          & Stopping condition & Recommendation & Objective\\
         \hline 
         $\tau_{\eps_1}$ & $Z_1^{\eps_1}(t) > 0 \; \land   Z_2^{\eps_1}(t) > 0$ & $\cO({\tau_{\varepsilon_1}})$ & $\eps_1$-PSI\\
         \hline 
         $\tau_{\eps_1,\eps_2}$ & $ Z_1^{\eps_1, \eps_2}(t) > 0 \; \land    Z_2^{\eps_1, \eps_2}(t) > 0$ & $\OPT^{\eps_1}(\tau_{\eps_1,\eps_2})$& $(\eps_1,\eps_2)$-PSI\\
         \hline  
         $\tau^{k}$ & $\lvert \OPT^{\eps_1}(t)\lvert\geq k$ & $\OPT^{\eps_1}(\tau^{k})$ & $\eps_1$-PSI-$k$ \\\hline 
    \end{tabular}
    \caption{Stopping conditions and associated recommendation}
    \label{tab:stop-rec-rules}
\end{table}

\begin{restatable}{lemma}{correctnessGeneric}\label{lem:correctness-generic} Assume $\cE$ holds. For $\eps_1$-PSI (resp. $(\eps_1,\eps_2)$-PSI , $\eps_1$-PSI-$k$), Adaptive Pareto Exploration combined with the stopping rule $\tau_{\eps_1}$ (resp. $\tau_{\eps_1,\eps_2}$, resp. $\tau_{\eps_1}^{k}$) outputs a correct subset. 
\end{restatable}
\begin{remark}We decoupled the presentation of the sampling rule to that of the ``sequential testing'' aspect (stopping and recommendation). We could even go further and observe that multiple tests could actually be run in parallel, for free. If we collect samples with APE (which only depends on $\eps_1$), whenever one of the three stopping conditions given in Table~\ref{tab:stop-rec-rules} triggers,  \emph{for any values of $\varepsilon_2$ or $k$}, we can decide to stop 
and make the corresponding recommendation or continue and wait for another ``more interesting'' stopping condition to be satisfied. If $\cE$ holds, a recommendation made at any such time will be correct for the objective associated to the stopping criterion (third column in Table~\ref{tab:stop-rec-rules}).     
\end{remark}

\subsection{Our instantiation} \label{subsec:concrete}
We propose to instantiate the algorithms with confidence interval on the difference of pair of arms.
For any pair $i,j\in \bA$, we define a function $\beta_{i,j}$ such that for any $d\in [D]$,  $U^d_{i,j}(t) = \muh_i^d(t) - \muh^d_j(t) + \beta_{i,j}(t)$ and $L^d_{i,j}(t) = \muh_i^d(t) - \muh^d_j(t) - \beta_{i,j}(t)$. We take from  \cite{kaufmann_mixture_2021} the following confidence bonus for time-uniform concentration:
\begin{equation}
\label{eq:beta}
    \beta_{i,j}(t):= 2\sqrt{\left(C^g\left(\frac{\log\left(\frac{K_1}{\delta}\right)}{2}\right) + \sum_{a\in \{ i ,j\}}
    \log(4 + \log(T_a(t)))\right) \left(\sum_{a\in \{ i ,j\}}\frac{1}{T_a(t)}\right)},
\end{equation}
where $K_1 := K(K-1)D/2$ and $C^g \approx x + \log(x)$ is a calibration function. They result in the simple expressions  $\Mh^{\pm}(i,j,t) = \Mh(i,j,t) \pm \beta_{i,j}(t)$ and $\mh^{\pm}(i,j,t) = \mh(i,j,t) \pm \beta_{i,j}(t)$. As an example, we state in Algorithm~\ref{alg:alg1} the pseudo-code of APE combined the stopping rule suited for the $k$-relaxation of $\varepsilon_1$-PSI, which we refer to as $\varepsilon_1$-APE-$k$. 

\begin{algorithm}[hbt]
\caption{$\varepsilon_1$-APE-$k$: Adaptive Pareto Exploration for $\varepsilon_1$-PSI-$k$}\label{alg:alg1}
\KwData{parameter $\eps_1 \geq 0$, $k \in [K]$}
\Input{ sample each arm once, set $t=K$, $T_i(K)=1$ for any $i\in \bA$} 
 \For{$t=K+1,\dots,$}{
  $S(t)= \left\{i \in \bA: \forall  j  \in \bA\backslash \{i\},  \Mh(i,j,t)>0 \right\}$\;
  $\OPT^{\eps_1}(t)=\left\{i \in \bA: \forall  j  \in \bA\backslash \{i\},  \Mh(i,j,t) - \beta_{i,j}(t) +\eps_1>0 \right\}$\;
   \If{$\lvert \OPT^{\eps_1}(t)\lvert\geq k$}{\textbf{break} and output $\OPT^{\eps_1}(t)$}
  \If{$Z_1^{\eps_1}(t) > 0 \; \land   Z_2^{\eps_1}(t) > 0$}{
  \textbf{break} and output $\cO({t}) = S(t) \bigcup \left\{ i \in S(t)^\complement: \nexists j\neq i: \mh(i,j,t)- \beta_{i,j}(t)> 0\right\}$
  }
  $b_t := \argmax_{i \in \bA \setminus \OPT^{\eps_1}(t)} \min_{j \neq i }\left[\Mh(i,j,t) + \beta_{i,j}(t)\right]$\;
     $c_t := \argmin_{i \neq b_t} \left[\Mh(b_t, j, t) - \beta_{b_t,j}(t)\right]$\;
  \Sample{$a_t := \argmin_{i \in \{b_t, c_t\}} T_i(t)$}\;
   }
\end{algorithm}


In Appendix~\ref{sec:another_instanciation}, we also study a different instantiation based on confidence bounds of the form $U_{i,j}(t) = U_i(t) - L_j(t)$ where $[L_i(t),U_i(t)]$ is a confidence interval on $\mu_i$. This is the approach followed by LUCB for $D=1$ and prior work on Pareto identification \cite{auer_pareto_2016,zuluaga_e-pal_2016}. In practice we advocate the use of the pairwise confidence intervals defined above, even if our current analysis does not allow to quantify their improvement. For the LUCB-like instantiation, we further derive in Appendix~\ref{sec:another_instanciation} an upper bound on the expected stopping time of \algname for the different stopping rules. 

\section{Theoretical analysis} \label{sec:analysis}
In this section, we state our main theorem on the sample complexity of our algorithms and give a sketch of its proof. 

First let us introduce some quantities that are needed to state the theorem. 
The sample complexity of the algorithm proposed by \cite{auer_pareto_2016} for $(\eps_1)$-Pareto set identification scales as a sum over the arms $i$ of $1/(\Delta_{i}\vee \varepsilon_1)^{2}$ where $\Delta_i$ is called the sub-optimality gap of arm $i$ and is defined as follows. For a sub-optimal arm $i\notin \cS^\star(\cX)$, 
\begin{align*}
    \Delta_i &:= \max_{j \in \cS^\star} \m(i,j), 
\end{align*}
which is the smallest quantity that should be added component-wise to $\vmu_i$ to make $i$ appear Pareto optimal w.r.t $\{\vmu_i: i\in \bA \}$.
For a Pareto optimal arm $i \in \cS^\star(\cX)$, the definition is more involved: 
\begin{align*}
    \Delta_i := \begin{cases}
   \min_{j\in \bA \setminus \{i\}} \Delta_j  \quad &\tm{if }\cS^\star:=\{i\}\\
    \min (\delta_i^+, \delta_i^-)\quad &\tm{else,} 
 \end{cases}
\end{align*}
where $$\delta_i^+:= \min_{j\in \cS^\star \setminus \{i\}} \min(\M(i,j), \M(j,i)) \ \text{and} \ \ \ \delta_i^-:= \min_{j\in \bA\setminus \cS^\star}\{(\M(j,i))^+ + \Delta_{j}\}.$$ For $x\in \bR, (x)^+:= \max(x, 0)$.
We also introduce some additional notion needed to express the contribution of the $k$-relaxation. Let $1\leq k\leq K$. 
For any arm $i$, let $\omega_i = \min_{j\neq i } \M(i,j)$ and define 
$$ \omega^k:=  \overset{k}{\max\limits_{i \in \bA} } \; \omega_i, \qquad \cS^{\star, k} := \overset{1\dots k}{\argmax\limits_{i \in \bA} } \; \omega_i, $$
with the $k$-th max and first to $k$-th argmax  operators.  Observe that $\omega^k > 0$ if and only if $|\cS^\star(\cX)| \geq k$. 

\begin{restatable}{theorem}{mainTheorem}\label{thm:sampcompl}
Fix a risk parameter $\delta \in (0, 1)$, $\eps_1\geq 0$, let $k\leq K$ and $\nu$ a bandit with $1$-subgaussian marginals. With probability at least $1 - \delta$, 
$\eps_1$-{APE}-$k$ recommends a correct set for the $\eps_1$-{PSI}-$k$ objective and stops after at most 
\begin{equation*}
    \sum_{a \in \bA} \frac{88}{{\widetilde\Delta_a}^2}\log\left(\frac{2K(K-1)D}{\delta} \log\left(\frac{12e}{{\widetilde\Delta_a}}\right)\right), 
\end{equation*}
samples, where for each $a\in \bA$, $\widetilde{\Delta}_a := \max(\Delta_a, \eps_1,\omega^k)$.
\end{restatable}

First, when $k=K$,  observing that $\eps_1$-APE-$K$ provides a $\delta$-correct algorithm for $\varepsilon_1$-PSI, our bound improves the result of \cite{auer_pareto_2016} for the $\eps_1$-PSI problem in terms of constant multiplicative factors and $\log\log\Delta^{-1}$ terms instead of $\log\Delta^{-2}$. It nearly matches the lower bound of Auer et al.\cite{auer_pareto_2016} for the $\eps_1$-PSI problem (Theorem 17 therein). 
It also shows the impact of the $k$-relaxation on the sample complexity. In particular, we can remark that for any arm $i\in \cS^\star \backslash \cS^{\star, k}$, $\max(\Delta_i, \omega_k) = \omega_k$.  Intuitively, it says that we shouldn't pay more than the cost of identifying the $k$-th optimal arm, ordered by the $\omega_i'$s. A similar result has been obtained for the \emph{any $k-$sized subset of the best $m$} problem \cite{chaudhuri_pac_nodate}. But the authors have shown the relaxation only for the best $m$ arms while our result shows that even the sub-optimal arms should be sampled less.

\revision{In Appendix~\ref{sec:lower_bounds}, we prove the following lower bound showing that in some scenarios, $\eps_1$-APE-$k$ is optimal for $\varepsilon_1$-PSI-$k$ , up to $D\log(K)$ and constant multiplicative terms. We note that for $\varepsilon_1$-PSI a 
lower bound featuring the gaps $\Delta_a$ and $\varepsilon_1$ was already derived by Auer et al. \cite{auer_pareto_2016}.

\begin{restatable}{theorem}{lowerBound}
  There exists a bandit instance $\nu$ with $\lvert \cS^\star \lvert = p\geq 3$  such that for $k \in \{ p, p-1, p-2\} $ any $\delta$-correct algorithm for $0$-PSI-$k$ verifies
    \[ \bE_\nu(\tau_\delta) \geq \frac{1}{D} \log\left(\frac1\delta\right) \sum_{a=1}^K\frac{1}{(\Delta_a^k)^2},\]
    where $\Delta_a^k := \Delta_a + \omega^k$ and $\tau_\delta$ is the stopping time of the algorithm. 
\end{restatable}}

In Appendix~\ref{sec:alg_eps_2}, we prove that \autoref{thm:sampcompl} without the $\omega^k$ terms also holds for $(\eps_1, \eps_2)$-APE. This does not justifies the reduction in sample complexity when setting $\varepsilon_2>0$ in $(\varepsilon_1,\varepsilon_2)$-PSI observed in our experiments but it at least guarantees that the $\varepsilon_2$-relaxation doesn't make things worse.   

Furthermore, since our algorithm allows $\eps_1=0$, it is  also an algorithm for BAI when $D=1, \eps_1=0$. We prove in Appendix~\ref{sec:bai} that in this case, the gaps $\Delta_i's$ matches  the classical gaps in BAI \cite{audibert_best_2010, kaufmann_complexity_2014} and we derive its sample complexity from \autoref{thm:sampcompl} showing that it is similar in theory to UGap \cite{gabillon_best_2012}, LUCB\cite{kalyanakrishnan_pac_2012} and LUCB++ \cite{simchowitz_simulator_2017} but have better empirical performance. 

\paragraph{Sketch of proof of Theorem~\ref{thm:sampcompl}} 
Using Proposition 24 of \cite{kaufmann_mixture_2021} we first prove that the choice of $\beta_{i,j}$ in \eqref{eq:beta} yields $\bP(\cE) \geq 1-\delta$ for the good event $\cE$ defined in~\eqref{eq:def_good_event}. Combining this result with \autoref{lem:correctness-generic}, yields that $\eps_1$-APE-$k$ is correct with probability at least $1-\delta$. 

\revision{The idea of the remaining proof is to show that under the event $\cE$, if \algname has not stopped at the end of round $t$, then the selected arm $a_t$ has not been explored enough. The first lemma showing this is specific to the stopping rule $\tau_{\varepsilon_1}^{k}$ used for $\varepsilon_1$-PSI-$k$.

\begin{restatable}{lemma}{samplwm}\label{lem:samplwm} Let $\eps_1\geq$ and $k\in [K]$. If  $\cE_t$ holds and $t<\tau_{\eps_1}^{k}$ then $\omega^k \leq 2 \beta_{a_t, a_t}(t)$. 
\end{restatable}

The next two lemmas are more general as they apply to different stopping rules.

\begin{restatable}{lemma}{samplold}\label{lem:samplold}  Let $\varepsilon_1 \geq 0$. Let $\tau = \tau_{\eps_1}^{k}$ for some $k \in [K]$ or $\tau = \tau_{\varepsilon_1,\varepsilon_2}$ for some $\varepsilon_2 \geq 0$. 
If $\cE_t$ holds and $t<\tau$ then $\Delta_{a_t} \leq 2 \beta_{a_t, a_t}(t)$.  
\end{restatable}

\begin{restatable}{lemma}{samplComplexEps}\label{lem:sampl-complex-eps}  
  Let $\varepsilon_1\geq 0$ and $\tau$ be as in Lemma~\ref{lem:samplold}. If $\cE_t$ holds and $t < \tau$ then $\eps_1 \leq 2 \beta_{a_t, a_t}(t)$. 
\end{restatable} 

As can be seen in Appendix~\ref{sec:proof_thm1}, the proofs of these three lemmas heavily rely on the specific definition of $b_t$ and $c_t$. In particular, to prove Lemma~\ref{lem:samplold} and \ref{lem:sampl-complex-eps}, we first establish that when $t < \tau$, any arm $j\in \bA$ satisfies $\mh(b_t,j,t)\leq \beta_{b_t, j}(t).$ The most sophisticated proof is then that of Lemma~\ref{lem:samplold}, which relies on a case distinction based on whether $b_t$ or $c_t$ belongs to the set of optimal arms.

These lemmas permit to show that, on $\cE_t$  if $t < \tau_{\eps_1}^{k}$ then $\widetilde{\Delta}_{a_t} < 2 \beta_{a_t,a_t}(t) \leq 2 \beta^{T_{a_t}(t)}$, where we define $\beta^n$ to be the expression of $\beta_{i,j}(t)$ when $T_i(t)=T_j(t) = n$. Then we have
\begin{eqnarray}
    \tau_{\eps_1}^k\ind\{\cE\} & \leq & \sum_{t=1}^\infty \sum_{a \in \bA} \ind{\left\{ \{a_t = a\} \land \left\{\widetilde \Delta_{a} \leq 2 \beta_{a}^{T_a(t)}\right\}  \right\}}\nonumber\\
    &\leq& \sum_{a \in \bA} \inf\left\{ n\geq 2 : \widetilde\Delta_a > 2 \beta^n\right\}\;.\label{eq:bounding-inverting-tau}
\end{eqnarray}
A careful upper-bounding of the RHS of \eqref{eq:bounding-inverting-tau} completes the proof of \autoref{thm:sampcompl}. }

\qed

\section{Experiments}
\label{sec:num_sim}
We evaluate the performance of Adaptive Pareto Exploration on a real-world scenario and on synthetic random Bernoulli instances. 
For a fair comparison, Algorithm 1 of \cite{auer_pareto_2016}, referred to as \algauername and \algname are both run with our confidence bonuses $\beta_{i,j}(t)$ on pairs of arms, which considerably improve single-arm confidence bonuses\footnote{In their experiments, \cite{auer_pareto_2016} already proposed the heuristic use of confidence bonuses of this form}. As anytime confidence bounds are known to be conservative, we use $K_1=1$ in \eqref{eq:beta} instead of its theoretical value coming from a union bound. Still, in all our experiments, the empirical error probability was (significantly) smaller than the target $\delta=0.1$. 
\paragraph{Real-world dataset} COV-BOOST \cite{munro_safety_2021} is phase 2 trial which was conducted on 2883 participants to measure the immunogenicity of different Covid-19 vaccines as third dose (booster) in various combinations of initially received vaccines (first two doses). This resulted in a total of 20 vaccination strategies being assessed, each of them defined by the vaccines received as first, second and third dose. The authors have reported the average responses induced by each candidate strategy on cohorts of participants, measuring several immunogenicity markers.
From this study, we extract and process the average response of each strategy to 3 specific immunogenicity indicators: two markers of antibody response and one of the cellular response. The outcomes are assumed to have a log-normal distribution \cite{munro_safety_2021}. We use the average (log) outcomes and their variances to simulate a multivariate Gaussian bandit with $K=20, D=3$. 
We give in Appendix~\ref{subsec:data_processing} some additional details about the processing of the data, and report the means and variances of each arm. In Appendix~\ref{subsec:implementation} we further explain how APE can be simply adapted when the marginals distributions of the arms have different variances.  
 \begin{figure}[htb]
     \centering
     \begin{subfigure}[b]{0.49\textwidth}
         \centering
    \includegraphics[width=\textwidth]{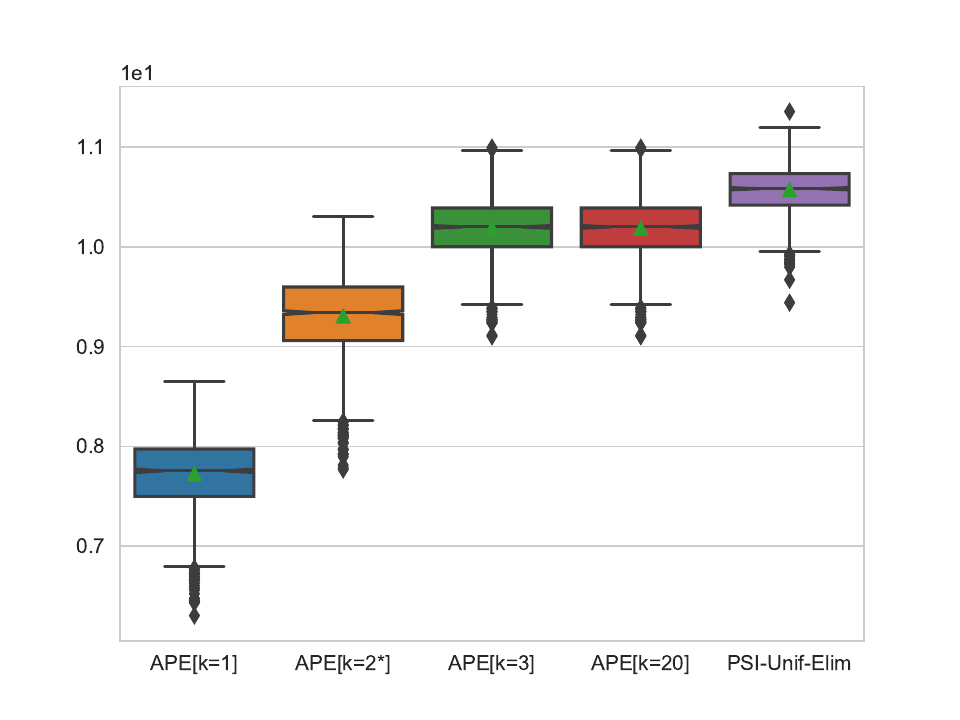}
         \label{fig:result-instanceI}
     \end{subfigure}
     \hfill
     \begin{subfigure}[b]{0.49\textwidth}
         \centering
\includegraphics[width=\textwidth]{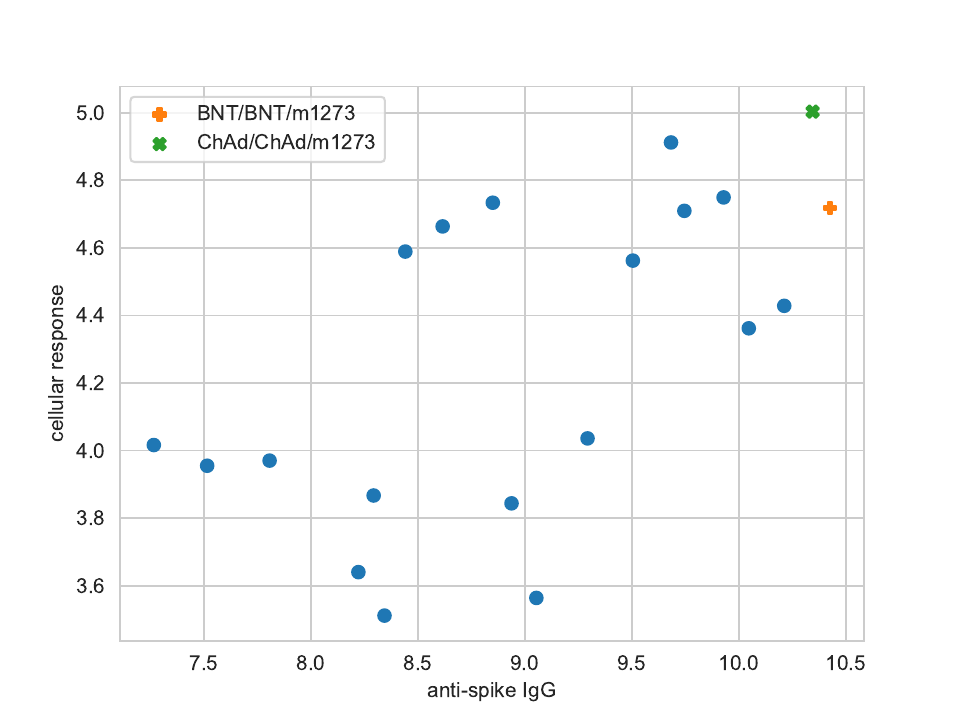}
         \label{fig:result-instanceII}
     \end{subfigure}     
        \caption{On the left is the log of the empirical sample complexity of \algauername and \algname on the real-world scenario plot (right) for 2 out of the 3 immunogenicity indicators.} 
        \label{fig:main-real-world}
\end{figure}
In this experiment, we set $\eps_1=0, \delta=0.1$ and compare \algauername to $0$-APE-$k$ (called APE-$k$ in the sequel) for different values of $k$. The empirical distribution of the sample complexity of the algorithms, averaged over 2000 independent runs, are reported in \autoref{fig:main-real-world}. The results are shown in log-scale (\texttt{y}-axis is the log of the sample complexity) to fit in the same figure. As $|\cS^\star|=2$, we first observe that, without the relaxation (i.e. for $k>3$), APE outperforms its state-of-the-art competitor \algauername. Moreover for $k=1$ or $k=2$, the sample complexity of APE-$k$ is significantly reduced. For $k=2$ when the stopping time $\tau^k$ is reached some sub-optimal arms have possibly not yet been identified as such, while for $k=3$, even if the optimal arms have been identified, the remaining arms have to be sampled enough to ensure that they are sub-optimal before stopping. This explains the gap in sample complexity between $k=2$ and $k=3$. 
In Appendix~\ref{subsec:additional_experiments}, we compare \algname to an adaptation of \algauername for the $k$-relaxation, showing that APE is always preferable.
%
\paragraph{Experiments on random instances}  To further illustrate the impact of the $k$-relaxation and to strengthen the comparison with \algauername, we ran the previous algorithms on 2000 randomly generated multi-variate Bernoulli instances, with $K=5$ arms and different values of the dimension $D$. We set $\delta =0.1$ and $\eps_1=0.005$ (to have reasonable running time). The averaged sample complexities are reported in \autoref{tab:bayes-commun}. 
We observe that APE (with $k=K$) uses $20$ to $25\%$ less samples than \algauername and tends to be more efficient as the dimension increases (and likely the size of the {Pareto set}, since the instances are randomly generated). We also note that identifying a $k$-sized subset of the {Pareto set} requires considerably less samples than exact PSI. In Appendix~\ref{subsec:additional_experiments} we also provide examples of instances for which APE takes up to 3 times less samples than \algauername.
\begin{table}[htb]
    \centering
    \begin{tabular}{c|c|c|c|c|c|c}
    \hline 
        & $\eps_1$-APE-$1$ & $\eps_1$-APE-$2$ &$\eps_1$-APE-$3$& $\eps_1$-APE-$4$ & $\eps_1$-APE-$5$ & $\eps_1$-\algauername \\
      \hline    
      $D=2$ & 811 & 39530 & 109020 &145777 & 150844 & 190625\\
      $D=4$ & 214&  6410 & 19908  & 68061 & 124001 & 157584\\
       $D=8$ &119&  204  &  405 & 1448    & 20336 & 27270\\
       \hline 
    \end{tabular}
    \caption{Average sample complexity over 2000 random Bernoulli instances with $K=5$ arms. On average the size of the Pareto set was $(2.295, 4.0625, 4.931)$ respectively for the dimensions $2, 4, 8$.}
    \label{tab:bayes-commun}
\end{table}

\begin{figure}[htb]
     \centering
     \begin{subfigure}[b]{0.49\textwidth}
         \centering
\includegraphics[width=\textwidth]{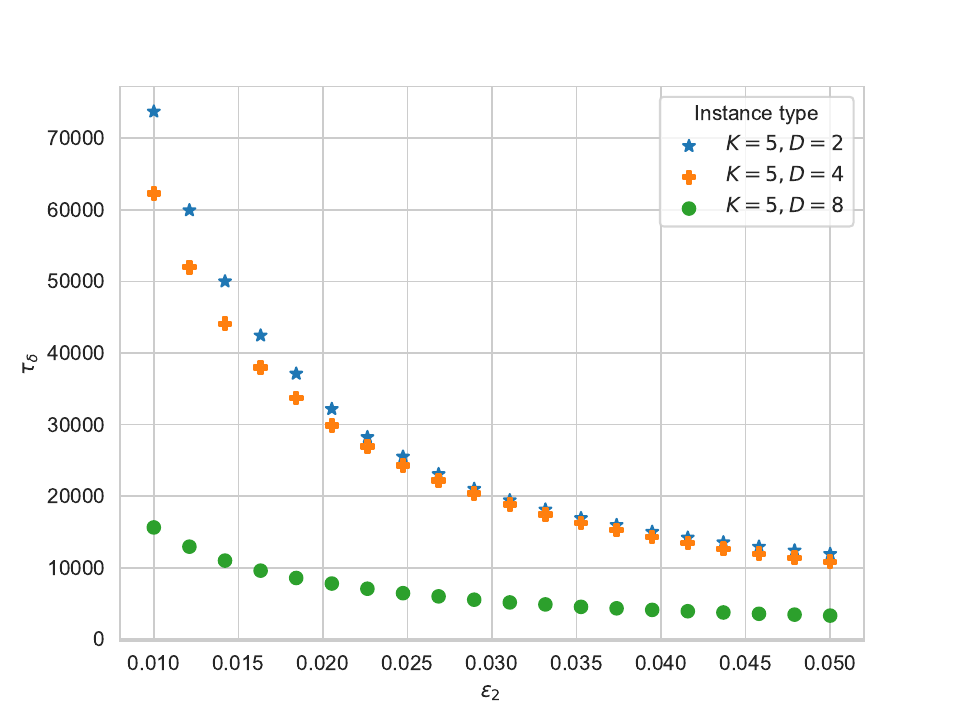}
         \caption{Average sample complexity vs $\eps_2$}
         \label{fig:main-sample_complexity_vs_eps_2}
     \end{subfigure}
     \hfill
     \begin{subfigure}[b]{0.49\textwidth}
         \centering
    \includegraphics[width=\textwidth]{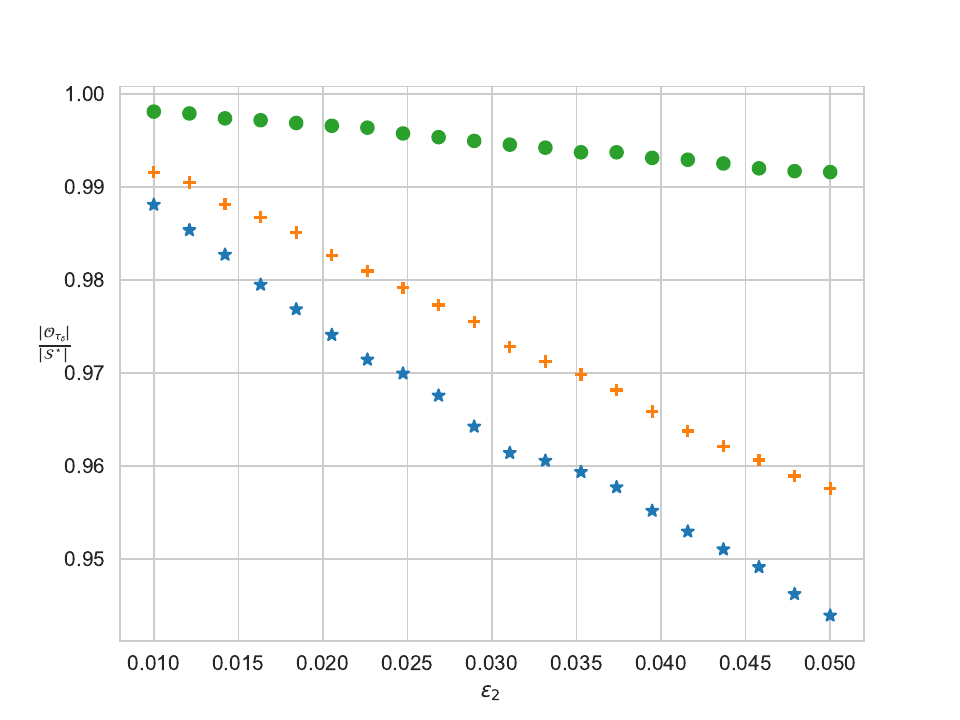}
         \caption{Average ratio $\nicefrac{\lvert \cO({\tau_{0,\eps_2}})\lvert}{\lvert \cS^\star\lvert }$ vs $\eps_2$}
         \label{fig:main-average_card_O}
     \end{subfigure}
        \caption{APE with $\tau_{0,\varepsilon_2}$ averaged over 2000 random Bernoulli instance with $K=5$ arms.}
        \label{fig:main-results-eps2}
\end{figure}

To illustrate the impact of the $\varepsilon_2$ relaxation, setting $\eps_1=0$ we report the sample complexity of \algname associated with the stopping time $\tau_{0, \eps_2}$ for 20 equally spaced values of $\eps_2 \in [0.01,0.05]$, averaged over 2000 random Bernoulli instances. \autoref{fig:main-results-eps2} shows the decrease of the average sample complexity when $\eps_2$ increases (left) and the average ratio of the size of the returned set to the size of the {Pareto set} (right). Note that for $\eps_1=0$, we have $\cO({\tau_{0, \eps_2}}) \subset \cS^\star$. The average sample complexity reported decreases up to $86\%$ for the instance with $K=5, D=2$ while the returned set contains more than $90\%$ of the Pareto optimal arms. In Appendix~\ref{subsec:additional_experiments}, we further illustrate the behavior of APE with the $\varepsilon_2$ relaxation on a fixed instance in dimension 2.


\section{Conclusion and perspective }

We proposed and analyzed APE, an adaptive sampling rule in multi-variate bandits models that when coupled with different stopping rules can 
tackle different relaxations of the fixed-confidence Pareto Set Identification problem. Our experiments revealed the good performance of the resulting algorithms compared to the state-of-the-art PSI algorithm as well as the great reductions in sample complexity brought by the relaxations. 
In future work, we intend to make our algorithms more practical for possible applications to clinical trials. For this purpose, as measuring efficacy takes time, we will investigate its adaptation to a batch setting, following, e.g. the work of \cite{jun_top_2016} for BAI. We will also investigate the use of APE beyond the fixed-confidence setting, to the possibly more realistic fixed-budget \cite{audibert_best_2010} or anytime exploration \cite{jun_anytime_2016} settings. To the best of our knowledge, no algorithm exists in the literature for PSI in such settings. Finally, following the works of \cite{zuluaga_active_2013,auer_pareto_2016}, we defined the $\varepsilon_1,\varepsilon_2$ relaxations with scalar values, so that the same slack applies to all components.   
Although we could easily modify our algorithms to tackle vectorial values $\veps_1, \veps_2$, so far we could only prove a dependence on $\min_d\eps_1^d$ in the sample complexity. We intend to study the right quantification in the sample complexity when $\veps_1$ and $\veps_2$ are vectorial. 



\begin{ack}
Cyrille Kone is funded by an Inria/Inserm PhD grant. Emilie Kaufmann acknoweldges the support of the French National Research Agency under the projects BOLD (ANR-19-CE23-0026-04) and FATE (ANR22-CE23-0016-01).
\end{ack}


\bibliographystyle{plain}
\bibliography{main}

\newpage
\tableofcontents
\newpage
\appendix
\section*{Outline and Notation}

In this section, we provide an outline of the supplemental material and define some additional notation. \autoref{sec:correctness} proves the correctness of ours algorithms and some concentration lemmas.  In \autoref{sec:proof_thm1}, we prove \autoref{thm:sampcompl} and the lemmas used in its proof. 
In \autoref{sec:alg_eps_2} we analyze the correctness and sample complexity of \algname associated to the stopping time $\tau_{\eps_1, \eps_2}$. \autoref{sec:lower_bounds} describes our worst-case lower bound and in \autoref{sec:bai} we relate our algorithm to other algorithms for BAI. In \autoref{sec:another_instanciation} we derive an upper-bound on the expectation of the sample complexity of $\eps_1$-APE-$k$ with a LUCB1-like instantiation and 
in \autoref{sec:tech_lemmas} we recall or prove some technical lemmas that are used in the main proofs. Finally, 
in \autoref{sec:impl_exp} we give further details about the experiments together with additional experimental results. 
\begin{table}[ht]
    \centering
    \begin{tabular}{p{2cm}p{3cm}p{7cm}}
    \hline 
         Notation&Type&Description  \\
         \hline 
         $\bA$ & & Set of arms\\
         $K$ & $\bN^\star$ & Number of arms \\
         $D$ & $\bN^\star$ & Dimension or number of attributes \\ 
         $[n]$ & & $\{1,\dots, n\}$\\ 
         $\m$ & $\bA^2 \rightarrow \bR$ & $\m(i,j) := \min\{ \mu_j^d - \mu_i^d: d\in [D] \}$\\
         $\M$ & $\bA^2 \rightarrow \bR$ & $\M(i,j) := \max\{ \mu_i^d - \mu_j^d: d\in [D] \}$ \\
         $(x)^+$ & $\bR\rightarrow \bR^+$ &$ \max(0, x)$ \\
         $\vmu_a$ & $\bR^D$ & Mean of arm $a \in \bA$ \\
         $\cS^\star_{\eps}$  & & $\{i \in \bA: \nexists j\neq i \in \bA : \vmu_i + \eps \prec \vmu_j \}$\\
         \hline 
    \end{tabular}
    \caption{Table of notation.}
    \label{tab:notation}
\end{table}

\section{Correctness for different stopping rules} 
\label{sec:correctness}
In this section, we gather and prove results that are related to the correctness of our algorithms, either for their generic form (\autoref{lem:ineq_gene} and \autoref{lem:correctness-generic}) or some specific calibration. We recall the definition of the events 
\[ \cE_t = \bigcap_{i=1}^K \bigcap_{j\neq i}\bigcap_{d=1}^D   \left\{L^d_{i,j}(t) \leq \mu_i^d - \mu_j^d \leq U^d_{i,j}(t)\right\}\ \text{ and } \ \cE = \bigcap_{t =1}^{\infty} \cE_{t}\;.\]

\subsection{Proof of \autoref{lem:ineq_gene}}
\ineqGene* 


\begin{proof}[Proof of \autoref{lem:ineq_gene}]
This result simply follows from the definition of $\cE_t$. Since
\begin{eqnarray*}
    \cE_t := \bigcap_{i=1}^K \bigcap_{j\neq i}\bigcap_{d=1}^D   \left\{ L^d_{i,j}(t) \leq \mu_i^d - \mu_j^d \leq U^d_{i,j}(t)\right \},
\end{eqnarray*}
if $\cE_t$ holds, then for any $i,j$
$$ \M^-(i,j,t):=\max_d L^d_{i,j}(t)\leq \M(i,j):= \max_d (\mu_i^d - \mu_j^d) \leq \max_d U^d_{i,j}(t):= \Mh^+(i,j,t),$$
and the second point follows by noting that $\m(i,j) =-\M(i,j)$ and $\mh^+(i,j,t) :=-\Mh^-(i,j,t); \mh^-(i,j,t) :=-\Mh^+(i,j,t)$ for any pair of arms.
\end{proof}
We remark that when the algorithm uses confidence bonus of form $ (\muh^d_i(t) - \muh^d_j(t)) \pm \beta_{i,j}(t)$,
\begin{eqnarray*}
 \Mh^+(i,j,t) &:=& \max_d U^d_{i,j}(t) = \max_d (\muh^d_i(t) - \muh^d_j(t)) + \beta_{i,j}(t) = \Mh(i,j,t) + \beta_{i,j}(t), \\  \Mh^-(i,j,t) &:=& \max_d L^d_{i,j}(t) = \max_d (\muh^d_i(t) - \muh^d_j(t)) - \beta_{i,j}(t) = \Mh(i,j,t) - \beta_{i,j}(t),
\end{eqnarray*}
and the previous lemma implies that on $\cE_t$, 
    \begin{eqnarray*}
        \lvert \M(i,j) - \Mh(i,j,t)\lvert \leq  \beta_{i,j}(t) \quad \text{and}\quad  \lvert \m(i,j) - \mh(i,j,t)\lvert \leq  \beta_{i,j}(t), 
    \end{eqnarray*}
    which is extensively used in our sample complexity analyses. 

%

\subsection{Proof of \autoref{lem:correctness-generic}}
\correctnessGeneric*

We show the correctness of $\eps_1$-PSI-$k$ (for any $k$) and we derive the correctness for $\eps_1$-PSI which is equivalent to $\eps_1$-PSI-$K$. The correctness of $(\eps_1, \eps_2)$-PSI is shown separately in \autoref{lem:correctness-eps1-eps-2-cover} (see \autoref{sec:alg_eps_2}). 
\begin{proof}[Proof of \autoref{lem:correctness-generic}]
  Assume $\cE$ holds. Let  $t=\tau_{\eps_1}^k$ and $i\in \OPT^{\eps_1}({t})$. Since $i \in \OPT^{\eps_1}(t)$, for any $j\neq i$, $$ \Mh(i,j) + \eps_1 \overset{\cE}{\geq}\Mh^-(i,j,t) + \eps_1 > 0,$$ that is $i \in \cS^\star_{\eps_1}$. Therefore, on the event $\cE$, $\OPT^{\eps_1}({t})\subset \cS^\star_{\eps_1}$. 
  Thus, if the stopping has occurred because $\lvert \OPT^{\eps_1}({t}) \lvert \geq k$, since in this case $\cO(t) \subset \OPT^{\eps_1}({t}) \subset \cS^\star_{\eps_1}$, all the recommended arms will be $(\eps_1)$-Pareto optimal. On the contrary, if $\lvert \OPT^{\eps_1}({t})\lvert < k$, then from the definition of $\tau_{\eps_1}^k$ it holds that  $$Z_1^{\eps_1}(t) > 0 \quad \text{and}\quad  Z_2^{\eps_1}(t) >0, $$
and the recommended set is then 
$$ \cO({t}) := 
    S(t) \cup \{ i \in S(t)^\complement: \nexists j\neq i: \mh^-(i,j,t)> 0\}. $$
For any $i\in \cO(t)^\complement$, by the definition of the recommended set and since $Z_2(t)>0$, $$\exists j \in \bA\; \tm{ such that } \m(i,j) \overset{\cE}{\geq}\mh^-(i,j,t)>0,$$ so $i$ is a sub-optimal arm. Therefore, $$\cS^\star \subset \cO(t).$$ 
Moreover, for any  $i \in \cO(t) \cap S(t)$,  since $Z_1^{\eps_1}(t)>0$ we have $h_i^{\eps_1}(t) >0$, that is 
\begin{equation}
\label{eq:eq-St}
\min_{j \in \bA \setminus\{ i\}} \M(i,j) + \eps_1 \overset{\cE}{\geq} \min_{j \in \bA\setminus \{ i\}} \Mh^-(i,j,t) + \eps_1 > 0.    
\end{equation}
If $i \in \cO(t) \cap S(t)^\complement$, by definition of $\cO(t)$, we have $g_i(t)<0$. However, since $Z_2^{\eps_1}(t)>0$,  $\max(g_i(t), h_i^{\eps_1}(t))>0$ so we also have $h_i^{\eps_1}(t)>0$ and \eqref{eq:eq-St} applies. Thus, for any $i\in \cO(t),$
$$\min_{j \in \bA \setminus\{ i\}} \M(i,j) + \eps_1 > 0,$$
that is $i\in \cS^\star_{\eps_1}$, so $\cS^\star \subset \cO(t) \subset \cS^\star_{\eps_1}$. Finally we can conclude that $\eps_1$-APE-$k$ and $\eps_1$-APE output a correct subset on $\cE$. 
\end{proof}

\subsection{Calibration of the confidence intervals}

In Section~\ref{subsec:concrete} we proposed to instantiate our algorithms with the confidence intervals 
\begin{equation}U^d_{i,j}(t) = \muh_i^d(t) - \muh^d_j(t) + \beta_{i,j}(t) \ \ \text{and} \ \ L^d_{i,j}(t) = \muh_i^d(t) - \muh^d_j(t) - \beta_{i,j}(t)\;.\label{eq:re-CI}\end{equation}
We prove below that $\cE$ is indeed a high-probability event for a suitable choice of $\beta_{i,j}(t)$.

\begin{lemma}
    \label{lem:prob-cE}
    Let $\nu$ be a bandit with $1$-subgaussian marginals. For the confidence intervals defined in \eqref{eq:re-CI}, with \[\beta_{i,j}(t)= 2\sqrt{\left(C^g\left(\frac{\log\left(\frac{K_1}{\delta}\right)}{2}\right) + \sum_{a\in \{ i ,j\}} \log(4 + \log(T_a(t)))\right) \left(\sum_{a\in \{ i ,j\}}\frac{1}{T_a(t)}\right)}.\]
 the event 
\begin{eqnarray*}
\cE = \bigcap_{t =1}^{\infty} \cE_{t}\; \ \text{ with } \ \ \cE_t = \bigcap_{i=1}^K \bigcap_{j\neq i}\bigcap_{d=1}^D   \left\{L^d_{i,j}(t) \leq \mu_i^d - \mu_j^d \leq U^d_{i,j}(t)\right\}
\end{eqnarray*} is such that  $\bP(\cE)\geq 1-\delta$.
\end{lemma}

\begin{proof}
By observing that for any pair of arm $\beta_{i,j} = \beta_{j,i}$, $\cE_t$ can be rewritten as 
\begin{eqnarray}
\cE_t &=& \bigcap_{\{i,j\} \in \Gamma} \bigcap_{d=1}^D   \left\{L^d_{i,j}(t, \delta) \leq \mu_i^d - \mu_j^d \leq U^d_{i,j}(t, \delta)\right\},     \\
&=&  \bigcap_{\{i,j\} \in \Gamma} \bigcap_{d=1}^D \left\{ \lvert (\muh^d_i(t) - \muh_j^d(t)) -(\mu_i^d - \mu_j^d) \lvert \leq \beta_{i,j}(t)\right\}, 
\end{eqnarray}
\end{proof}
where $\Gamma:= \binom{2}{[K]}$ is  the set of pair of 2 elements of $[K]$, which satisfies $\lvert \Gamma \lvert = K(K-1)/2$. Therefore, using a union bound, 
\begin{eqnarray*}
 \bP(\cE^\complement) &=& \bP(\exists t\geq 1: \cE_t^\complement \text{ holds }), \\
 &=& \bP\left(\exists t\geq 1, d\in [D], \{i,j\} \in \Gamma: \lvert (\muh^d_i(t) - \muh_j^d(t)) - (\mu_i^d - \mu_j^d) \lvert > \beta_{i,j}(t)\right), \\
 &\leq&  \sum_{\{i,j\}\in \Gamma} \sum_{d=1}^D \bP\left(\exists t\geq 1: \lvert (\muh^d_i(t) - \muh_j^d(t)) - (\mu_i^d - \mu_j^d) \lvert > \beta_{i,j}(t)\right), \\
 &\leq& \sum_{\{i,j\}\in \Gamma} \sum_{d=1}^D \frac{\delta}{K_1}  \quad \text{(by Proposition 24 of \cite{kaufmann_mixture_2021} which we recall below in \autoref{lem:prop24})},
\\&=& \delta,
\end{eqnarray*}
since $K_1:= \frac{K(K-1)D}{2}$ and $\lvert \Gamma \lvert = K(K-1)/2$. 

\begin{lemma}[Proposition 24 of \cite{kaufmann_mixture_2021}]
\label{lem:prop24}
Let $X, Y$ be centered $1-$subgaussian random variables and $\delta \in (0,1)$. 
Let $X, X_1, X_2, \dots$ be \iid  random variables and $Y, Y_1, Y_2, \dots$ be \iid random variables. 
With probability at least $1-\delta$, for all $p,q \geq 1$, 
    \[ \left\lvert \frac{1}{p}\sum_{s=1}^p X_s  - \frac{1}{q}\sum_{s=1}^q Y_s \right\lvert\leq 2\sqrt{\left(C^g\left(\frac{\log({1}/{\delta})}{2}\right) + \log\log(e^4 p) + \log\log(e^4q)\right) \left(\frac{1}{p} + \frac{1}{q}\right)} \]
where $C^g(x) \approx x + \log(x)$. 
\end{lemma}

\section{Sample complexity analysis}
\label{sec:proof_thm1}
In this section we prove \autoref{thm:sampcompl} which is restated below. 

\mainTheorem*

\begin{proof}[Proof of \autoref{thm:sampcompl}]The correctness follows from \autoref{lem:correctness-generic} and the fact that $\bP(\cE)\geq 1- \delta$ (\autoref{lem:prob-cE}). 
The upper-bound on the sample complexity is a direct consequence of \autoref{lem:samplwm}, \autoref{lem:samplold},  \autoref{lem:sampl-complex-eps} which are proved later in this section. 
    Indeed, using these lemmas we have that, if $\eps_1$-\algname-$k$ has not stopped during round $t$ i.e $t<\tau_{\eps_1}^k$ and the event $\cE_t$ holds, then 
    \begin{enumerate}[a)]
        \item $\omega^k\leq 2\beta_{a_t, a_t}(t)$ (\autoref{lem:samplwm}), 
        \item $\Delta_{a_t}\leq 2\beta_{a_t, a_t}(t)$(\autoref{lem:samplold}),
              \item $\eps_1 \leq 2\beta_{a_t, a_t}(t)$ (\autoref{lem:sampl-complex-eps})
    \end{enumerate}
    hold simultaneously. Then, if we do not count the first $K$ rounds due to initialization, and letting $\widetilde \Delta_{a}:=\max(\omega^k, \eps_1, \Delta_{a})$, 
    \begin{eqnarray*}
        \tau_{\eps_1}^k \ind\{\cE\} - 1 &\leq& \sum_{t=1}^{\infty} \ind\{\cE\}\ind\{\tau_{\eps_1}^k>t\}, \\&\leq& \sum_{t=1}^\infty \ind{\{\max(\omega^k, \eps_1, \Delta_{a_t}) \leq 2 \beta_{a_t, a_t}(t)\}}\\
        &=& \sum_{t=1}^\infty \ind{\{\widetilde \Delta_{a_t} \leq 2 \beta_{a_t, a_t}(t)\}}\\
        &=& \sum_{t=1}^\infty \sum_{a=1}^K\ind{\{ \{a_t = a\} \land  \{\widetilde \Delta_{a}\leq 2 \beta_{a, a}(t)\} \}}\\
        &=&  \sum_{a=1}^K \sum_{t=1}^\infty \ind{\{ \{a_t = a\} \land \{\widetilde \Delta_{a} \leq 2 \beta_{a, a}(t)\}  \}} \\
        &\leq& \sum_{a=1}^K \inf\{n\geq 2: \widetilde \Delta_a > 2 \beta^n\}, 
    \end{eqnarray*}
where $\beta^n$ is the expression of $\beta_{i,j}(t)$ when $T_i(t)=T_j(t) = n$, that is 
\[\beta^n= 2\sqrt{\left(C^g\left(\frac{\log\left(\frac{K_1}{\delta}\right)}{2}\right) +2 \log(4 + \log(n))\right)\frac{2}{n}}\;.\]
Then, an inversion result given in \autoref{lem:bound_inv_beta_ek} yields 
$$ \inf\{s\geq 2 : 2\beta^s < \widetilde \Delta_a \} \leq \frac{88}{\widetilde \Delta_a^2}\log\left(\frac{2K(K-1)D}{\delta}\log\left(\frac{12e}{\widetilde \Delta_a}\right)\right).$$
Therefore, 
\begin{equation*}
   \tau_{\eps_1}^k \ind\{\cE\} \leq \sum_{a\in \bA}\frac{88}{\widetilde \Delta_a^2}\log\left(\frac{2K(K-1)D}{\delta}\log\left(\frac{12e}{\widetilde \Delta_a}\right)\right). 
\end{equation*}
\end{proof}
We will now prove the lemmas involved in the proof of the main theorem. Two of them (\autoref{lem:samplold},  \autoref{lem:sampl-complex-eps}) rely on the following result, which is an important consequence of the definition of the APE sampling rule.  

\revision{\begin{lemma}
\label{lem:stoppingcond} Let $\varepsilon_1 \geq 0, \varepsilon_2\geq 0$ and $k\in [K]$. If $\tau = \tau_{\varepsilon_1}^{k}$ or $\tau = \tau_{\varepsilon_1,\varepsilon_2}$ the following holds. If $t<\tau$  then for any $j\in \bA$, $\mh(b_t,j,t)\leq \beta_{b_t, j}(t).$
    \end{lemma}}
    
\begin{proof}

The proof is split in two steps. 
\item 
    \paragraph{Step 1} If $t< \tau_{\eps_1}^k$ then for any $j\in \bA$, $\mh(b_t,j,t)\leq \beta_{b_t, j}(t)$.
    
First, note that $t<\max(\tau_{\eps_1}^k, \tau_{\eps_1})$ implies that $Z_1^{\eps_1}(t) \leq 0 $ or $Z_2^{\eps_1}(t) \leq 0$.
By definition of $b_t$ and noting that $\M(i,j,t) = -\m(i,j,t)$, we have 
\begin{equation}
\label{eq:equiv-bt}
 b_t \in \argmin_{i\in \OPT^{\eps_1}(t)^\complement} \max_{j\neq i }\mh(i,j,t) - \beta_{i,j}(t).   
\end{equation}
so that if there exists $j$ such that $\mh(b_t,j,t)>\beta_{b_t, j}(t)$, then 
$$ \max_{j\neq b_t } \mh(b_t, j,t) - \beta_{b_t, j}(t) >0, $$
therefore, 
\begin{equation}
 \forall i \in \OPT^{\eps_1}(t)^\complement\;, \max_{j\neq i}\mh(i,j,t) - \beta_{i,j}(t) >0 \quad \text{i.e} \quad g_i(t) >0. 
\end{equation}
    Furthermore, for any $i \in \OPT^{\eps_1}(t)$, $h^{\eps_1}_i(t) >0$. Putting things together, if there exists $j$ such that $\mh(b_t, j,t)>\beta_{b_t, j}(t)$ then, $Z^{\eps_1}_1(t) > 0$ and $Z^{\eps_1}_2(t)> 0$. 
\item 
\paragraph{Step 2}  If $t<\tau_{\eps_1, \eps_2}$ then for any $j\in \bA$, $\mh(b_t,j,t)\leq \beta_{b_t, j}(t)$. 

Recall that by definition $t<\tau_{\eps_1, \eps_2}$ implies that 
       $ Z_1^{\eps_1, \eps_2}(t) \leq 0 $ or $ Z_2^{\eps_1, \eps_2}(t) \leq 0$. Using \eqref{eq:equiv-bt}, if there exists $j$ such that $\mh(b_t,j,t)>\beta_{b_t, j}(t)$, then 
$$ \max_{j\neq b_t } \mh(b_t, j,t) - \beta_{b_t, j}(t) >0.$$
Combining this with 
$$g_i^{\eps_2}(t) := \max_{j \in \bA \setminus\{i\}}  \mh(i,j,t) -\beta_{i,j}(t) + \eps_2 \ind\{j \in \OPT^{\eps_1}(t)\}, $$
yields 
\begin{equation}
 \forall i \in \OPT^{\eps_1}(t)^\complement\;, 0<\max_{j\neq i}\mh(i,j,t) - \beta_{i,j}(t)\leq g^{\eps_2}_i(t).   
\end{equation}
Furthermore, since we have 
\begin{equation}
    \forall i \in \OPT^{\eps_1}(t)\;, h^{\eps_1}_i(t) >0, 
\end{equation}
the initial assumption would yield that for any arm $i$, $\max(h^{\eps_1}_i(t), g^{\eps_2}_i(t))>0$, so $Z_1^{\eps_1, \eps_2}(t)>0$ and $Z_2^{\eps_1, \eps_2}(t)>0$. 

\revision{We conclude that if $\tau = \tau_{\varepsilon_1}^{k}$ or $\tau_{\varepsilon_1,\varepsilon_2}$, $t<\tau$ implies that 
for any $j\in \bA, \mh(b_t,j,t)\leq \beta_{b_t, j}(t)$. }    
   \end{proof}

\subsection{Proof of Lemma~\ref{lem:samplwm}}
\samplwm*
\begin{proof}[Proof of \autoref{lem:samplwm}]
First, note that if $k>\lvert \cS^\star \lvert$, then the lemma holds trivially since $\omega_k <0$. In the sequel, we assume  $\cE_t$ holds and $k\leq\lvert\cS^\star\lvert$. If $t < \tau_{\eps_1}^k$ then it holds that $\lvert\OPT^{\eps_1}(t)\lvert < k$. So $\cS^{\star, k} \cap {\OPT^{\eps_1}(t)}^\complement \neq \emptyset$. Let $i\in\cS^{\star, k} \cap {\OPT^{\eps_1}(t)}^\complement$, we have 
\begin{eqnarray*}
    \omega^k &\leq& \omega_i = \min_{j \in \bA \setminus\{i\}} \M(i,j),\\
    &\overset{(a)}{\leq}& \min_{j \in \bA \setminus\{i\}} \M(i,j, t) + \beta_{i,j}(t), \\
    &\overset{(b)}{\leq}& \min_{j \in \bA \setminus\{b_t\}} \Mh(b_t,j, t) + \beta_{b_t, j}(t),\\
    &\leq& \Mh(b_t,c_t, t) + \beta_{b_t, c_t}(t) ,\\
    &{\overset{(c)}{\leq}}& \ 2\beta_{b_t, c_t}(t),\\
    &\leq& 2\beta_{a_t, a_t}(t),
\end{eqnarray*}
where $(a)$ uses that $\cE_t$ holds and \autoref{lem:ineq_gene}, $(b)$ uses the definition of $b_t$ and $(c)$ follows from the definition of $c_t$ and the fact that $b_t \notin \OPT^{\eps
_1}(t)$,which yields  $\Mh(b_t, c_t, t)\leq \beta_{b_t, c_t}(t)$. The last inequality follows since $a_t$ is the least sampled among $b_t, c_t$ and $\beta$ is decreasing. 
\end{proof}

\subsection{Proof of Lemma~\ref{lem:samplold}}
\samplold*
Before proving the \autoref{lem:samplold}, we state the following lemma which is used to derive an upper bound on the gap of an optimal arm. Its proof is postponed to the end of the section. 
\begin{lemma}
\label{lem:gap_opt}
    For any Pareto optimal arm $i$, $\Delta_i \leq \min_{j\neq i} \M(i,j).$
\end{lemma}
\begin{proof}[Proof of \autoref{lem:samplold}]
Assume that $\cE_t$ holds. We consider four different cases depending on whether $b_t$ and $c_t$ are optimal or sub-optimal. 
\item 
\paragraph{Case 1.1}$b_t$ is a Pareto optimal arm. 
From the definition of the gap of an optimal arm and using \autoref{lem:gap_opt} it follows 
$\Delta_{b_t} \leq \Mh(b_t, c_t)$ which on $\cE_t$ and using \autoref{lem:ineq_gene} yields 
\begin{equation}
  \Delta_{b_t}  + \eps_1 \leq \Mh(b_t,c_t, t) + \beta_{b_t, c_t}(t) + \eps_1 
\end{equation}
then, noting that there exists $j \in \bA\setminus\{ b_t\}$ such that $\Mh(b_t, j,t) + \eps_1 \leq \beta_{b_t, j}(t)$, by definition of $c_t$, we have 
\begin{equation}
 \Mh(b_t, c_t, t)+ \eps_1 \leq \beta_{b_t, c_t}(t),
\end{equation}
therefore, 
\begin{eqnarray*}
    \Delta_{b_t} + \eps_1 &\leq&  2\beta_{b_t, c_t}(t).
\end{eqnarray*}
\item 
\paragraph{Case 1.2}$b_t$ is a sub-optimal arm. By definition of $c_t$ and using $\M=-\m$, we have 
\begin{align}
 c_t &\in \argmax_{j\in \bA \setminus\{ b_t\}}  \mh(b_t, j, t) + \beta_{b_t, j}(t), 
\end{align}
then, from the definition of the gap of a sub-optimal arm and since $\cE_t$ holds, we know that there exists an arm $b_t^\star$ such that 
\begin{eqnarray*}
    \Delta_{b_t} = \m(b_t, b_t^\star) &\leq&\mh(b_t, b_t^\star, t) + \beta_{b_t, b_t^\star}(t),\\
    &\overset{(a)}{\leq}& \mh(b_t, c_t, t) + \beta_{b_t, c_t}(t), \\
    &\overset{(b)}{\leq}& 2\beta_{b_t, c_t}(t).
\end{eqnarray*}
where $(a)$ uses the definition of $c_t$ and $(b)$ uses Lemma~\ref{lem:stoppingcond}.
\item 
\paragraph{Case 2.1} $c_t$ is a Pareto optimal arm. If $b_t$ is also an optimal arm,  it follows that $\Delta_{c_t}\leq \Mh(b_t, c_t)$ which on $\cE_t$ yields $\Delta_{c_t}\leq \Mh(b_t, c_t, t) + \beta_{b_t, c_t}(t)$, then, similarly to case 1.1, we have $\Mh(b_t, j,t) + \eps_1 \leq \beta_{b_t, j}(t)$  so $$\Delta_{c_t} + \eps_1 {\leq} 2 \beta_{b_t, c_t}(t).$$ Now, assume $b_t$ is a sub-optimal arm. Then, by definition, $\Delta_{c_t} \leq \M(b_t, c_t)^+ + \Delta_{b_t}$. Using a similar reasoning to case 1.2, it holds that $\Delta_{b_t}\leq \mh(b_t, c_t, t) + \beta_{b_t, c_t}(t)$, so
\begin{eqnarray*}
    \Delta_{c_t}&\leq& \M(b_t, c_t)^+ + \Delta_{b_t},\\
    &\leq& (\Mh(b_t, c_t, t) + \beta_{b_t, c_t}(t))^+  + \mh(b_t, c_t,t)+ \beta_{b_t, c_t}(t),\\
    &=& (-\mh(b_t, c_t, t) + \beta_{b_t, c_t}(t) )^+  + \mh(b_t, c_t,t)+ \beta_{b_t, c_t}(t), \\
    &\overset{(a)}{\leq}& \max(2\beta_{b_t, c_t}(t), \mh(b_t, c_t,t)+ \beta_{b_t, c_t}(t)) \\
    &\overset{(b)}{\leq}& 2\beta_{b_t, c_t}(t).
\end{eqnarray*}
where $(a)$ follows from $(x-y)^+  + (x + y) \leq \max(x+y, 2x)$ and $(b)$ follows from $\mh(b_t, c_t, t)\leq \beta_{b_t, c_t}(t)$ (\autoref{lem:stoppingcond}). 
\item 
\paragraph{Case 2.2} $c_t$ is a sub-optimal arm. We know that there exists an arm $c_t^\star$ such that $\Delta_{c_t} = \m(c_t, c_t^\star)$. If $c_t^\star = b_t$ then, since $\m(j,i) \leq \M(i,j)$ (follows from the definition), we have 
\begin{eqnarray*}
    \Delta_{c_t} =\m(c_t, c_t^\star) &=& \m(c_t, b_t),\\
    &\leq& \M(b_t, c_t), \\
    &\overset{(a)}{\leq}& \Mh(b_t, c_t, t) + \beta_{b_t, c_t}(t), \\
 &\overset{(b)}{\leq}& 2\beta_{b_t, c_t}(t), 
\end{eqnarray*}
where $(a)$ follows from $\cE_t$ and $(b)$ has been already justified in the case 1.1. If  $b_t\neq c_t^\star$, then by definition of $c_t$, we have 
$$ \mh(b_t, c_t, t) + \beta_{b_t, c_t}(t) \geq \mh(b_t, c_t^\star, t) + \beta_{b_t, c_t^\star}(t),$$
which implies that there exists $d\in [D]$ such that 
$$ \muh_{c_t}^d(t) - \muh_{b_t}^d(t) + \beta_{b_t, c_t}(t) \geq \muh_{c_t^\star}^d(t) - \muh_{b_t}^d(t) + \beta_{b_t, c_t^\star}(t) \overset{\cE_t}{\geq} \mu_{c_t^\star}^d - \mu_{b_t}^d,$$
then recalling that $\beta_{i,j} = \beta_{j,i}$, 
$$ \mu_{c_t}^d - \mu_{b_t}^d + 2\beta_{b_t, c_t}(t) \overset{\cE_t}{\geq} (\muh_{c_t}^d(t) - \muh_{b_t}^d(t) - \beta_{b_t, c_t}(t)) + 2\beta_{b_t, c_t}(t) \geq \mu_{c_t^\star}^d - \mu_{b_t}^d.$$
Put together, there exists $d\in [D]$ such that 
\begin{eqnarray*}
\mu_{c_t^\star}^d - \mu_{c_t}^d \leq  2\beta_{b_t, c_t}(t),
\end{eqnarray*}
so $$ \Delta_{c_t} = \min_d (\mu_{c_t^\star}^d - \mu_{c_t}^d) \leq 2\beta_{b_t, c_t}(t),$$
Putting the four cases together, we have proved that if $t<\max(\tau_{\eps_1}^k, \tau_{\eps_1, \eps_2})$ then both 
\begin{equation}
\label{eq:eq-lem-samplod}
  \Delta_{b_t} \leq 2 \beta_{b_t, c_t}(t) \quad \text{and}\quad  \Delta_{c_t} \leq 2\beta_{b_t, c_t}(t)
\end{equation}
holds. Further noting that $a_t$ is the least sampled among among $b_t, c_t$ and $\beta$ is non-increasing, $\beta_{b_t, c_t}(t) \leq \beta_{a_t, a_t}(t)$, \eqref{eq:eq-lem-samplod} yields 
$$ \Delta_{a_t} \leq 2\beta_{a_t, a_t}(t),$$
which achieves the proof. 
\end{proof}
\subsection{Proof of \autoref{lem:sampl-complex-eps}}
The following lemma holds for each of the stopping times $\tau_{\eps_1}, \tau_{\eps_1, \eps_2}$ and $\tau_{\eps_1}^k$. 
\samplComplexEps*
\begin{proof}[Proof of \autoref{lem:sampl-complex-eps}]
By \autoref{lem:stoppingcond}, we have $\mh(b_t, c_t,t) \leq \beta_{b_t, c_t}(t)$ or equivalently \begin{equation}\Mh(b_t,c_t,t) \geq  - \beta_{b_t, c_t}(t)\;.\label{eq:step1}\end{equation} Then, knowing that $b_t \notin \OPT^{\eps_1}(t)$, there exists an arm $j$ such that $ \eps_1 + \Mh(b_t,j,t) \leq \beta_{b_t, j}(t)$. Using further the definition of $c_t$, it follows that $\eps_1 + M(b_t,c_t,t) \leq \beta_{b_t, c_t}(t)$. Combining this with inequality \eqref{eq:step1} and noting that $a_t$ is the least sampled among $b_t, c_t$ yields $$\beta_{a_t, a_t}(t) \geq \beta_{b_t, c_t}(t) \geq \eps_1/2.$$ 
\end{proof}

\subsection{Auxiliary results} 

We state the following lemma which is used to prove \autoref{lem:gap_opt}. 
\begin{lemma} 
\label{lem:existence_dom}
For any sub-optimal arm $a$, there exists a Pareto optimal arm $a^\star$ such that $\vmu_a \prec \vmu_{a^\star}$ and $\Delta_{a} = \m(a, a^\star)>0$. Moreover, For any $i\in \bA\setminus \cS^\star$, $j \in \cS^\star$, 
\begin{enumerate}[i)]
    \item $\max_{k\in \cS^\star} \m(i,k) = \max_{k\in \bA} \m(i,k)$, 
    \item If $i \in \argmin_{k\in \bA \setminus \{j\}} \M(j,k)$ then $j$ is the unique arm such that $\vmu_i \prec \vmu_j$ 
\end{enumerate}
\end{lemma}
\begin{proof}
Assume there are $p<n$ dominated arms. Without loss of generality, we may assume they are $\vmu_1, \dots, \vmu_p$. Let $i_1\leq p$. Suppose that no Pareto-optimal arm dominates $\vmu_{i_1}$. Since $\vmu_{i_1}$ is not optimal, by the latter assumption, there exists $i_2 \leq p$ such that $\vmu_{i_1}\prec \vmu_{i_2}$. If $\vmu_{i_2}$ is dominated by a Pareto optimal arm, this arm also dominates $\vmu_{i_1}$ (strict dominance is transitive) which contradicts the initial assumption. If not, there exits $i_3 \leq p$ such that $\vmu_{i_1}\prec \vmu_{i_2}\prec \vmu_{i_3}$. Again we can use the same reasoning as before for $i_3$. In any case we should stop in at most $p$ steps, otherwise we would have $\vmu_{i_1}\prec \vmu_{i_2}\prec \dots \prec \vmu_{i_p}$ and $\vmu_{i_p}$ should be dominated by a Pareto-optimal arm, otherwise it would be itself Pareto-optimal, which is not the case. Therefore, for any $a\in \bA\setminus \cS^\star$, there exists $a^\star \in \cS^\star$ such that $a^\star \prec a \tm{ and } \Delta_a = \m(a, a^\star)>0$.

 Letting $i$ be a sub-optimal arm, since for any $a\in \bA\setminus \cS^\star$, there exists $a^\star \in \cS^\star$  such that $a\prec a^\star$, it follows that
\[ \forall  d \in [D],\; \mu_a^d - \mu_i^d < \mu_{a^\star}^d - \mu_i^d,\]
which leads to $\m(i, a) \leq \m(i, a^\star)$, so 
\[  \max_{j\in \bA} \m(i,j) = \max_{j\in \cS^\star} \m(i,j) > 0,\]
which achieves the proof of the first point {\it i)}. For the second point, let $q \in \bA\setminus \cS^\star$ and $q'$ such that $q\prec q'$ and \[ q \in \argmin_{a\in \bA \setminus \{j\}} \M(j,a).\] By direct algebra, since $q\prec q'$, we have 
\[ \M(j, q') < \M(j, q),\] which is impossible if  $q'\neq j$ (because $q$ belongs to the  argmin). Therefore, if $$q \in \argmin_{a\in \bA \setminus \{j\}} \M(j,a)$$ is a sub-optimal arm, then $j$ is the only arm such that $q\prec j$ (i.e $\vmu_q \prec \vmu_j$). 
\end{proof}
We now prove \autoref{lem:gap_opt} which follows  from the previous lemma. 
\begin{proof}[Proof of \autoref{lem:gap_opt}]
If $\argmin_{j\neq i} \M(i,j) \subset \cS^\star$, then the lemma follows from the definition of the gap of an optimal arm recalled in Section 4. If $\min_{j\neq i} \M(i,j) = \M(i, a)$, $a\notin \cS^\star$, then, 
    from \autoref{lem:existence_dom}, $i$ is the unique arm which dominates $a$ so $\Delta_a = \m(a, i)$ and using the definition of the gap of an optimal arm, \begin{eqnarray*}
        \Delta_i &\leq& \M(a, i)^+ + \Delta_a, \\
        &=& 0 + \m(a, i) \leq \M(i,a),
    \end{eqnarray*}
    where we have used the the fact that  $\m(p,q)\leq \M(q, p)$  for any pair of arms $p,q$ (which follows from the definition). Therefore, for an optimal arm $i$, we always have 
    $$ \Delta_i \leq \min_{j\neq i} \M(i,j).$$
\end{proof}

\section{Algorithm for finding an $(\eps_1,\eps_2)$-cover}
\label{sec:alg_eps_2}
In this section, we analyse the sample complexity of \algname when it is associated to the stopping time  $\tau_{\eps_1, \eps_2}$ for identifying an $(\eps_1, \eps_2)$-cover of the {Pareto set}.  The sampling rule remains unchanged an we prove that the algorithm does not require more samples to find an $(\eps_1, \eps_2)$-cover than to solve the $\eps_1$-PSI problem. 

\begin{algorithm}[H]
\caption{$(\eps_1, \eps_2)$-\algname}\label{alg:alg-eps-2}
\KwData{parameter $\eps_1\geq 0$}
\Input{sample each arm once, set $t=K$, $T_i(t)=1$ for any $i\in \bA$} 
 \For{$t=K+1,\dots,$}{
  $b_t := \argmax_{i \in \bA \setminus \OPT^{\eps_1}(t)} \min_{j \neq i }\Mh^+(i,j,t)$\;
     $c_t := \argmin_{j\neq b_t} \Mh^{-}(b_t, j, t)$\;
     \If{$Z_1^{\eps_1, \eps_2}(t) > 0 \land Z_2^{\eps_1, \eps_2}(t)>0$}{
     \textbf{break} and output $\OPT^{\eps_1}(t)$\;
     }
  \Sample{$a_t := \argmin_{i \in \{b_t, c_t\}} T_i(t)$}\;
 }
\end{algorithm}
We recall the stopping time $\tau_{\eps_1, \eps_2}$. 
\paragraph{Stopping rule} Let $\eps_1, \eps_2\geq 0$ and  $0<\delta<1$. Then
\begin{equation}
\label{eq:stopping-rule}
	\tau_{\eps_1, \eps_2}:= \inf \left\{ t\geq K :\;  Z_1^{\eps_1, \eps_2}(t) > 0 \; \land    Z_2^{\eps_1, \eps_2}(t) > 0\right\},
\end{equation}
where,
\begin{eqnarray*}
 Z_1^{\eps_1, \eps_2}(t) &:=& \min_{i \in S(t)}  \max(g_i^{\eps_2}(t), h_i^{\eps_1}(t))\\
 Z_2^{\eps_1, \eps_2}(t) &:=& \min_{i \in S(t)^{\complement}}  \max(g_i^{\eps_2}(t), h_i^{\eps_1}(t)),
\end{eqnarray*}
and
\begin{eqnarray*}
  g_i^{\eps_2}(t) &:=& \max_{j \in \bA \setminus\{i\}}  \mh^-(i,j,t)  + \eps_2 \ind\{j \in \OPT^{\eps_1}(t)\} 
  \\
  h_i^{\eps_1}(t) &:=& \min_{j\in\bA\setminus\{i\}}\Mh^-(i,j,t)+\eps_1 
\end{eqnarray*}
\paragraph{Recommendation rule}When 
it is associated to the stopping time $\tau_{\eps_1, \eps_2}$, \algname recommends
$$\cO({\tau_{\eps_1, \eps_2}}) := \OPT^{\eps_1}(\tau_{\eps_1, \eps_2})
,$$ which can be understood as follows. When $\tau_{\eps_1, \eps_2}$ is reached, the arms that are not yet identified as (nearly) optimal are either $\eps_2$-dominated by an arm in $\OPT^{\eps_1}(\tau_{\eps_1, \eps_2})$ or sub-optimal, which is proven formally in \autoref{lem:correctness-eps1-eps-2-cover}. 
\begin{lemma}
\label{lem:correctness-eps1-eps-2-cover}
    Fix $\delta \in (0, 1), \eps_1,\eps_2 \geq0$ then ${(\eps_1,\eps_2})$-\algname recommends an $(\eps_1, \eps_2)$-cover of the {Pareto set} on the event $\cE$.  \end{lemma}
    \begin{proof}[Proof of \autoref{lem:correctness-eps1-eps-2-cover}]
Assume $\cE$ holds. Let  $t=\tau_{\eps_1, \eps_2}$ and $i\in \OPT^{\eps_1}({t})$. Since $i\in \OPT^{\eps_1}({t})$, for any $j\neq i$, $ \M(i,j) + {\eps_1}\overset{\cE}{\geq}\Mh^-(i,j,t) + {\eps_1}> 0$ that is $i \in \cS^\star_{\eps_1}$. Therefore, on the event $\cE$, $\OPT^{\eps_1}({t})\subset \cS^\star_{\eps_1}$. When the stopping time $\tau_\delta^{\eps_1, \eps_2}$ is reached, $Z^{\eps_1, \eps_2}_1(t) > 0$ and $Z^{\eps_1, \eps_2}_2(t) >0$. 
 Under this condition, $$\OPT^{\eps_1}(t)\neq \emptyset.$$ Indeed, since $Z^{\eps_1, \eps_2}_1(t)>0$ and $Z^{\eps_1, \eps_2}_2(t)>0$, if $\OPT^{\eps_1}(t) = \emptyset$ then, by the stopping rule and since $\OPT^{\eps_1}(t) = \emptyset$, for any arm $i$, we would have $h_i^{\eps_1}(t)<0$  and $g_i^{\eps_2}(t)>0$. That is, for any arm $i \in \bA$, 
      $$\exists j\neq i \tm{ such that } \m(i,j)\overset{\cE}{>} \mh^-(i,j,t)>0,$$ so every arm would be strictly dominated, which is impossible since the {Pareto set} cannot be empty. Then, $\OPT^{\eps
_1}(t)\neq \emptyset$ and 
    for any $i\in \cO(t)^\complement
 = \OPT^{\eps_1}(t)^\complement$, by the stopping rule it holds that $\max(g_i^{\eps_2}(t), h_i^{\eps_1}(t)) > 0$. Further noting that for such arm $i\in \OPT^{\eps_1}(t)^\complement$, $h_i^{\eps_1}(t)<0$ , we thus have $g_i^{\eps_2}(t)>0$, that is $$ \mh^-(i,j,t)  +  \eps_2 \ind\{j \in \cO(t)\} >0,$$
 which on the event $\cE$ yields 
 $$ \m(i,j) + \eps_2 \ind\{j \in \cO(t)\}  >0.$$
 Therefore, for such arm $i$, either
    \begin{enumerate}[i)]
    \item $\exists j \in \bA \tm{ such that } \m(i,j)>0$ that is $\vmu_i \prec \vmu_j$ or 
    \item $\exists j \in \cO({t})$ 
    \tm{ such that } $\m(i,j) + \eps_2 > 0$ that is $\vmu_i \prec \vmu_j + \boldsymbol{\eps_2}$ with $\boldsymbol{\eps_2}:= (\eps_2, \dots, \eps_2)$. 
    \end{enumerate}
Put together, $\cO(t) \subset \cS^\star_{\eps_1}$ and  for any $i\notin \cO(t)$, either $i \notin \cS^\star$ ($i$ is a sub-optimal arm) or there exists $j\in \cO(t)$ such that $\vmu_i \prec \vmu_j + \boldsymbol{\eps_2}$. Thus $\cO(t)$ is an $(\eps_1, \eps_2)$-cover of the {Pareto set} and $(\eps_1, \eps_2)$-\algname is correct for $(\eps_1, \eps_2)$-cover identification. 
\end{proof}
The two lemmas restated below are used to prove identically to \autoref{thm:sampcompl}, the main theorem of this section. 
\samplold*
The following lemma holds for each of the stopping times $\tau_{\eps_1}, \tau_{\eps_1, \eps_2}$ and $\tau_{\eps_1}^k$. 
\samplComplexEps*
\begin{theorem}
\label{thm:sampcompl2}
Fix $\delta \in (0, 1)$, $\eps_1, \eps_2 \geq 0$. Then $(\eps_1, \eps_2)$-\algname outputs an $(\eps_1, \eps_2)$-cover of the {Pareto set} with probability at least  $1-\delta$ using at most 
\begin{equation}
     \sum_{a \in \bA} \frac{88}{(\Delta_a^{\eps})^2} \log\left(\frac{2K(K-1)D}\delta \log\left(\frac{12e}{\Delta^{\eps}_{a}}\right)\right)
\end{equation}
samples, where for all $a\in \bA$, $\Delta_a^{\eps} := \max(\Delta_a, \eps_1)$. 
\end{theorem}

This is the first problem-dependent sample complexity upper-bound for the $(\eps_1, \eps_2)$-cover of the {Pareto set}. In particular, this result holds for the $\eps$-accurate {Pareto set} identification \cite{zuluaga_e-pal_2016} which corresponds to the particular case $\eps_1 = \eps_2 = \eps$ of the {Pareto set} cover.  Therefore, $(\eps, \eps)$-\algname could be compared to $\eps$-PAL for $\eps$-accurate {Pareto set}, which however relies on a Gaussian process modeling assumption. 

While this sample complexity result upper-bound does not clearly show the dependence in $\eps_2$, we note that for some problems, we have a nearly matching lower bound that does not depend on $\eps_2$. In particular, consider the case $D=1, \eps_1 = 0, \eps_2 >0$ and assume there is a unique best arm (classical assumption in BAI) $a_\star$. For this setting, an algorithm for $(\eps_1, \eps_2)$-cover identification is required to output a set $\hat S$ such that $\hat S \subset \cS^\star = \{ a_\star\}$ and for any $i\neq a_\star$ either $\mu_i < \mu_{a_\star}$ or $\mu_i \leq \mu_{a_\star} + \eps_2$ which trivially holds as long as $\hat S \subset \cS^\star$. Therefore, this problem is equivalent to (exact) Best Arm Identification. 
    Almost matched lower bounds for BAI are known and does not depend on $\eps_2$ (\cite{kaufmann_complexity_2014, simchowitz_simulator_2017, garivier_optimal_2016}). 
This observation can be generalized to any configuration where there is a unique (Pareto) optimal arm. Letting $D\geq 1, \eps_1=0, \eps_2>0$ and $\nu$ a bandit with one Pareto optimal arm $a_\star$, any algorithm for $(\eps_1, \eps_2)$-covering is required to output a set $\hat S \subset \cS^\star = \{ a_\star\}$. And for any $i\neq a_\star$ either $\vmu_i \prec \vmu_{a_\star}$ or $\vmu_i \prec \vmu_{a_\star} + \eps_2$ which trivially holds as long as $\hat S \subset \cS^\star = \{ a_\star\}$. So, on theses instances, $(0, \eps_2)$-covering is equivalent $0$-PSI and the nearly matched lower of \cite{auer_pareto_2016} for $0$-PSI does not depend on $\eps_2$ (Theorem 17 therein). 

In our experiments (see \autoref{subsec:additional_experiments}), we will see that in configurations with multiple Pareto optimal arms, the parameter $\varepsilon_2$ can still help to empirically reduce the sample complexity. Quantifying its precise impact on the sample complexity is left as future work.

\section{Lower Bound}
\label{sec:lower_bounds}
In this section, we give a gap-dependent lower-bound for the $k$-relaxation in some configurations. We use the change of distribution lemma of \cite{kaufmann_complexity_2014} (Lemma 1 therein). 


\lowerBound*

 \begin{proof} Let $p = K = \lvert \cS^\star\lvert$. 
        w.l.o.g assume $\cS^\star = \{1, \dots, p\}$ and $\cS^{\star,k} = \{1, \dots, k\}$. 
        Let $\vmu_0 \in \bR^D$ and for $(p-2)\leq i\leq p$, define 
        $$ \mu_i^d := \begin{cases}
            \phantom{-}2^{p-i}\omega  &\tm{ if } $d = 1$\\
            -2^{p-i}\omega &\tm{ else if } $d = 2$ \\
            \phantom{-}\mu_0^d &  \tm{ else. }
        \end{cases},$$
        for $1\leq i\leq p-3$, 

        $$\mu_i^d := \begin{cases}
           \phantom{-}(4 +2i) \omega &\tm{ if } $d = 1$\\
             - (4 +2i)\omega &\tm{ else if } $d = 2$ \\
            \phantom{-}\mu_0^d &  \tm{ else. }
        \end{cases}$$
Let $\nu$ be a bandit where each arm $i$ is a multivariate Gaussian
with mean $\vmu_i$ and covariance matrix $I_D$ i.e $\nu_i \sim \cN(\vmu_i, I_D)$ (with $I_D$ the identity matrix in dimension $D$). By direct calculation, for $1\leq i,j\leq p -3$,
$$\M(i,j) = \M(j,i) = 2\omega\lvert i - j\lvert ,$$ and for $p-2\leq i, j\leq p$, 
$$\M(i, j) = \M(j,i) = 2^p\omega\lvert 2^{-i} - 2^{-j}\lvert,$$
for $i\leq p-3$ and $(p-2)\leq j\leq p$, 
$$\M(i, j) = \M(j, i) = (4 + 2i - 2^{p-j})\omega \geq 2\omega.$$
Therefore, computing $\omega_i$ and $\delta_i^+$ for any $i\in [p]$ yields
\begin{eqnarray*}
    \delta_i^+ &:=& \min_{j\in [p]\backslash \{i\}}\min(\M(i,j), \M(j,i)),
\\
&=&  \begin{cases}
    \omega & \tm{ if } i=p,\\
    2^{p-i-1} \omega & \tm{ if } i \in \{p-2, p-1\} \\
    2\omega & \tm{ else, } 
\end{cases} 
\end{eqnarray*}
additionally, for any $i\leq p$, 
\begin{eqnarray*}
 \omega_i &:=& \min_{j\neq i} \M(i,j), \\ &=&  \begin{cases}
    \omega & \tm{ if } i=p,\\
    2^{p-i-1} \omega & \tm{ if } i \in \{p-2, p-1\} \\
    2\omega & \tm{ else.} 
\end{cases}   
\end{eqnarray*}
Thus,  $$\omega^{(p)} =\omega^{(p-1)} =  \omega \tm{ and }\omega^{(p-2)} = 2\omega.$$
Let $\gamma>0$. For any optimal arm $i$, since $\M(i, i+1) = \M(i+1, i) = \delta_{i}^+$, the vector 
$$ \vmu_i + \delta_i^+ + \gamma $$
Pareto dominates $\vmu_{i+1}$ or $\vmu_{i-1}$ and $\vmu_i - \delta_i^+ - \gamma \prec \vmu_{i+1} \tm{ or } \vmu_{i-1}$. Moreover, it is easy to observe that for $k \in \{p-2, p-1, p\}$ and any $i \in [p]$,
$$\vmu_{i} + \delta_i^+ + \omega^{(k)} + \gamma$$
Pareto dominates 1 (if $k\in \{p,p-1\}$)  or 2 (if $k=p-2$) other optimal arms. Letting $k\in \{p-2, p-1, p\}$, for any $i\in [p]$, we define the alternative bandit $\nu^{(i)}$ which is also Gaussian with the same covariance matrix $I_D$ and means given by 
\begin{equation}
    \vmu^{(i)}_j = \begin{cases}
        \vmu_j & \tm{ if } j\neq i \\ \vmu_j - \delta_i^+ - \omega^{(k)} - \gamma & \tm{ if } j = i \tm{ and }\bP_\nu( j \in \hat S) \geq \frac12\\
         \vmu_j + \delta_i^+  +\omega^{(k)} + \gamma & \tm{ if } j = i \tm{ and } \bP_\nu( j \in \hat S) <\frac12.
    \end{cases}
\end{equation}
Therefore, since $\cA$ is $\delta-$correct, and by what precedes,
\begin{itemize}
    \item $\tm{ if }\bP_\nu(i \in \hat S) \geq \frac12 \tm{ then } \bP_{\nu^{(i)}}(i \in \hat S) \leq \delta$ and 
    \item $\tm{ if }\bP_\nu(i \in \hat S) <\frac12 \tm{ then } \bP_{\nu^{(i)}}(i\in \hat S) \geq 1-\delta$. 
\end{itemize}
The first point follows simply from the definition of $\delta_i^+$ and the fact that by design $\M(i,j) = \M(i, j)$ for $i,j\in [p],$ For the second point, if $k\in \{ p, p -1\}$, in the bandit $\nu^{(i)}$, at least one arm of $\cS^\star(\nu)$ is no longer optimal, then $\lvert \cS^\star(\nu^{(i)}) \lvert \leq p-1 \leq k.$ So $\bP_{\nu^{(i)}}(i\in \hat S)\geq 1-\delta$. If $k=p-2$ since two arms of $\cS^\star(\nu)$ are now dominated, we have $\lvert \cS^\star(\nu^{(i)})\lvert \leq p-2 = k,$ hence $\bP_{\nu^{(i)}}(i\in \hat S)\geq 1-\delta$
Letting $\KL$ denote the Kulback-Leiber divergence and using Lemma~1 of \cite{kaufmann_complexity_2014}, on $\cF_\tau-$measurable event $$E_i = \begin{cases}
    \{ i \in \hat S\} & \tm{ if } \bP_\nu(i\in \hat S)\geq \frac12, \\
    \{i \notin \hat S\} &\tm{ if } \bP_\nu(i\in \hat S)<\frac12,
\end{cases}$$
for which $\bP_\nu(E_i) \geq \frac12$ and $\bP_{\nu^{(i)}}(E_i)\leq \delta$, it comes that 
$$ \sum_{a\in \bA} \bE_\nu(T_a(\tau_\delta))\KL(\nu_a, \nu_a^{(i)}) \geq d(\bP_\nu(E_i), \bP_{\nu^{(i)}}(E_i)), $$
hence 
\begin{equation}
\label{eq:lemma_change_distrib}
 \bE_\nu(T_i(\tau_\delta))\KL(\nu_i, \nu_i^{(i)}) \geq d(\bP_\nu(E_i), \bP_{\nu^{(i)}}(E_i)),   
\end{equation}
where $d(x,y) = x\log(x/y) + (1-x)\log((1-x)/(1-y))$ is the binary relative entropy. Since $\bP_\nu(E_i)\geq \frac12$ and $\bP_{\nu^{(i)}}(E_i) \leq \delta$, 
\eqref{eq:lemma_change_distrib} yields (see \cite{kaufmann_complexity_2014}),
\begin{eqnarray*}
    \bE_\nu(T_i(\tau_\delta)) &\geq& \frac{1}{\KL(\nu_i, \nu_i^{(i)})} \frac12\left( \log\left(\frac{1}{2\delta}\right) + \log\left(\frac{1}{2(1-\delta)}\right)\right) \\
    &=& 
    \frac{1}{2\KL(\nu_i, \nu_i^{(i)})} \log\left( \frac{1}{\delta(1-\delta)}\right)\\
    &\geq& \frac{1}{2\KL(\nu_i, \nu_i^{(i)})} \log(1/\delta).
\end{eqnarray*}
By direct algebra, we compute (independent marginals since the covariance is diagonal $I_D$), $$\KL(\nu_i, \nu_i^{(i)}) = \frac12\| \vmu_j -\delta_i^+ - \omega^{(k)} - \gamma  -\vmu_j\|_2^2 = \frac D2(-\delta_i^+ - \omega^{(k)} - \gamma )^2.$$ Noting that on this instance all the arms are optimal, we have for any arm $i, \Delta_i = \delta_i^+$. Finally, letting $\gamma \longrightarrow 0$ proves that for any arm $i$, 
$$ \bE_\nu(T_i(\tau_\delta)) \geq \frac{1}{D(\Delta_i^k)^2}\log(1/\delta), $$ further noting that $\bE(\tau_\delta) = \sum_{i=1}^K \bE(T_i(\tau_\delta))$ 
achieves the proof. We have chosen a diagonal matrix matrix simplicity, we believe that choosing carefully correlated  objectives like in \cite{auer_pareto_2016} could give a tighter lower bound especially regarding the dependence in the dimension $D$. 
    \end{proof}

\section{Best Arm Identification}
\label{sec:bai}
In this section, we discuss the sample complexity and the performance of \algname associated to the stopping rule $\tau_{0}^1$ for BAI. Noting that when $D=1$, the {Pareto set} is just the \emph{argmax} over the means, BAI and PSI are the same for uni-dimensional bandits. For this setting we should expect algorithms for PSI to be competitive with existing algorithms for BAI. We will show that it is actually the case for \algname. Let $D=1$ and $\nu$ be a one-dimensional $K$-armed bandit. Letting $a_\star$ denote the unique optimal arm of the bandit $\nu$, i.e $\cS^\star =\{a_\star\}$, one can easily check that the gaps defined for PSI matches the common notion of gaps for BAI. Indeed, for any $a\neq a_\star$,
\begin{eqnarray*}
\Delta_{a}  &:=& \max_{j \in \cS^\star} \m(a, j), \\
&=& \m(a, a_\star) \\
&=&\mu_{a_\star} - \mu_a,
\end{eqnarray*}
and $$\Delta_{a_\star}  = \min_{j\neq a_\star} \{\M(j,a^\star)^+ + \Delta_j \} = \min_{j\neq a_\star} \Delta_j,$$
which matches the definition of the gap in the one-dimensional bandit setting (\cite{audibert_best_2010, kaufmann_complexity_2014}). Therefore, the sample complexity of \algname for BAI can be deduced from \autoref{thm:sampcompl}. 
\begin{theorem}
\label{thm:sampl_compl_BAI}
Let $\delta \in (0, 1), K\geq 2$ and $\nu$ a $K-$armed bandit with a unique best arm $a_\star$ and $1$-subgaussian distributions. \algname associated to be stopping time $\tau_0^1$ identifies the best arm $a^\star$ with probability at least $1-\delta$ using at most the following number of samples  \[ 
\sum_{a=1}^K \frac{88}{\Delta_a^2} \log\left(\frac{2K(K-1)}\delta \log\left(\frac{12e}{\Delta_{a}}\right)\right). 
\]
\end{theorem}
In particular, the $k$ relaxation is not meaningful in this setting. Under the unique optimal arm assumption, the algorithm will always stop when the best arm has been identified. And we remark that from the definition of $\omega_i$'s 
$$ \omega_1 = \min_{j\neq a_\star}\M(a_\star, j) = \min_{j\neq a_\star} \Delta_j = \Delta_{a_\star} \quad \text{ and } \quad \forall i\neq a_\star, \quad \omega_i < 0,$$
so for any $k\leq K$, $\max(\omega_k, \Delta_a) = \Delta_a$.
\begin{remark}
\label{rmk:improve_ape_for_BAI}
\autoref{thm:sampl_compl_BAI} could be slightly improved. On the event $\cE$ we consider that for any pair of arms the difference of their empirical mean does not deviate too much from its actual value. For BAI, since we know that there is a unique optimal arm (enforced by assumption), it is sufficient to control the difference between the best arm and any other arm, therefore we could replace the $K(K-1)/2$ term due to union bound in the confidence bonus by $K-1$ and we could show that this will reflect in the sample complexity by replacing $K(K-1)$ by $2(K-1)$. However, this cannot be done in general for PSI since we do not know in advance the number of optimal arms. 
\end{remark}
When $D=1$,  \algname reduces to sample at each round $t$, the least sampled among 
\begin{eqnarray}
\label{eq:sampling-rule-BAI}
    b_t &:=& \argmax_i \left\{\min_{j\neq i}U_{i,j}(t)\right\},\\
    c_t &:=& \argmin_{j\neq b_t} L_{b_t, j}(t),\label{eq:sampling-rule-BAI2}
\end{eqnarray}
where $U_{i,j}(t) := \muh_i(t) - \muh_j(t) + \beta_{i,j}(t)$ and $L_{i,j}(t):= \muh_i(t) - \muh_j(t) - \beta_{i,j}(t)$ are upper and lower bounds on  the difference $\mu_i - \mu_j$. To be in the same setting as LUCB and UGapEc which uses confidence interval on single arms, we would have $\beta_{i,j}(t) := \beta_{i}(t) + \beta_{j}(t)$, where $\beta_{i}'s$ are confidence bonuses on single arms such that $L_i(t) := \muh_i(t) - \beta_{i}(t)$ and $U_i(t):= \muh_i(t) + \beta_i(t)$ are lower and upper confidence bounds on $\mu_i$. Then \eqref{eq:sampling-rule-BAI} and \eqref{eq:sampling-rule-BAI2} rewrite as 
\begin{eqnarray*}
    b_t &:=& \argmax_i \left\{U_i(t) -\max_{j\neq i}L_{j}(t)\right\},\\
    c_t &:=& \argmax_{j\neq b_t} U_j(t),
\end{eqnarray*}
This resembles the sampling rule of UGap, which defines 
\begin{eqnarray*}
    b_t^{\text{UGap}} &:=& \argmax_{i} \left\{L_i(t)-\max_{j\neq i} U_j(t) \right\}, \\
    c_t^{\tm{UGap}} &:=& \argmax_{j\neq b_t} U_i(t),
\end{eqnarray*}

and also pulls the least sampled so far. We note that a variant of our algorithm in which both $b_t, c_t$ would be sampled (in the spirit of LUCB \cite{kalyanakrishnan_pac_2012}) could also be analyzed using the same arguments employed in the proof of \autoref{thm:sampcompl}. 

Note that when $\eps_1=0$, for any $i\in S(t)^\complement, g_i(t) > h_i^0(t)$. Indeed, by definition, $$h_i^0(t) = \min_{j\neq i} (\M(i,j,t) - \beta_{i,j}(t) ) = \min_{j\neq i} (-\mh(i,j,t) - \beta_{i,j}(t))$$ and since $i\in S(t)^\complement$, there exists $i^\star$ such that $\mh(i,i^\star,t)>0$ (i.e $\vmuh_i(t) \prec \vmuh_{i^\star}(t)$) and so 
$$ -\mh(i,i^\star,t) - \beta_{i,i^\star}(t) < \mh(i,i^\star,t) - \beta_{i,i^\star}(t).$$
Therefore, 
\begin{equation*}
 \min_{j\neq i} (-\mh(i,j,t) - \beta_{i,j}(t)):= h_i^0(t) < \max_{j\neq i} (\mh(i,j,t) - \beta_{i,j}(t)):= g_i(t).
\end{equation*}
Thus for $\eps_1=0$, 
\begin{equation}
 \label{eq:eq-simpli}   
 Z^0_2(t) = \min_{i\in S(t)^\complement} g_i(t). 
\end{equation}
In the sequel, for this section, we remove the dependence on $\eps_1$ to write $Z_i(t)$ instead of $Z_i^{0}$ for $i=0$ and $i=1$. 
In particular, when $D=1, \eps_1=0$, the stopping time $\tau_0$ can be simplified to 
$$ \tau_0 = \inf \{t\geq K : Z_1(t)>0\},$$
which is a consequence of the following lemma. 
\begin{lemma}
	\label{lem:stopping-time-simplif}
	For $D=1, \eps_1=0$, 
	$$ \inf \{t\in \bN^\star : Z_1(t)>0\} = \inf \{t\in \bN^\star : Z_1(t)>0 \land Z_2(t)>0\} .$$
\end{lemma}
\begin{proof}[Proof of \autoref{lem:stopping-time-simplif}]
 Let $S(t) = \{\hat a_t\}$. Using the definition of $h_i^0, g_i$ and \eqref{eq:eq-simpli}, $Z_1(t)$ and $Z_2(t)$ simplifies to  
         \begin{eqnarray}
             Z_1(t) &=&  \min_{i\neq \hat a_t} \left \{\muh_{\hat a_t}(t) - \muh_i(t) - \beta_{\hat a_t, i}(t)\right\}, 
         \\ Z_2(t) &=& \min_{i \neq \hat a_t} \left\{ \max_{j\neq i } \left[ \muh_j(t) - \muh_i(t) - \beta_{i,j}(t)\right]\right\}. 
         \end{eqnarray}
 We have : 
\begin{eqnarray*}
          Z_1(t) > 0 &\impl& \forall i\neq \hat a_t, \muh_{\hat a_t}(t) - \muh_i(t) - \beta_{\hat a_t, i}(t)>0,\\
          &\impl& \forall i\neq \hat a_t, \max_{j\neq i}\left[ \muh_{j}(t) - \muh_i(t) - \beta_{j, i}(t)\right]>0, \\
          &\impl& Z_2(t) >0. 
         \end{eqnarray*}
 Thus, $Z_1(t) > 0 \impl (Z_1(t)>0 \land Z_2(t)>0)$ and the reverse holds trivially. So 
 $$ Z_1(t) >0 \equi (Z_1(t)>0 \land Z_2(t)>0).$$

\end{proof}
Letting $\hat{a}_t$ denote the empirical best arm after $t$ rounds, the stopping rule of APE (with the instantiation proposed in Section~\ref{subsec:concrete} based on confidence intervals on pairs of arms) reduces to 
\[\tau_{0} = \inf \left\{t \geq K :  \forall i \neq \hat{a}_t, \frac{(\hat{\mu}_{\hat{a}_t}(t) - \hat{\mu}_i(t))^2}{2\left(\frac{1}{T_{\hat{a}_t}(t)} + \frac{1}{T_i(t)}\right)} \geq 2C^{g}\left(\frac{\log(K_1/\delta)}{2}\right) + 2\!\!\!\!\sum_{a \in \{\hat{a}_t,i\}} \!\!\!\log(4+\log( T_a(t)))\right\}\]
which is very close to a Generalized Likelihood Ratio (GLR) stopping rule assuming Gaussian distributions with variance 1 for the rewards (which is known to be also correct for sub-Gaussian rewards) \cite{garivier_optimal_2016,kaufmann_mixture_2021}. This modified stopping rule compared to LUCB1 and UGapEc can partially explains the empirical improvement observed in Section~\ref{subsec:additional_experiments}.



\section{LUCB1-like instantiation of \algname} 
\label{sec:another_instanciation}
In this section we derive an upper bound on the expectation of the sample complexity $\tau_{\eps_1}^k$ when \algname is run with confidence bonuses similar to LUCB1 \cite{kalyanakrishnan_pac_2012}. This is different from \autoref{thm:sampcompl} for which the sample complexity is bounded only on the high-probability event $\cE$ but as, for many algorithms in pure-exploration \cite{simchowitz_simulator_2017, gabillon_best_2012, auer_pareto_2016} we do not control what happens on $\cE^\complement$. Therefore, our goal here is to upper-bound $\bE(\tau)$ instead of $\bE(\ind\{\cE\}\tau)$ which we did in \autoref{thm:sampcompl}. To adapt the strategy employed in \cite{kalyanakrishnan_pac_2012}, we use similar confidence bonuses, thus we define for any arm $i$, 
\begin{equation}
    \label{eq:beta-single}
    \beta_i(t) = \sqrt{\frac{2}{T_i(t)}\log\left(\frac{5KDt^4}{2\delta}\right)},
\end{equation}
and for any pair $i,j\in\bA$, $\beta_{i,j}(t) = \beta_i(t) + \beta_j(t)$. Recalling the definition of $\cE$ and $\cE_t$ introduced in Section 3.1, 
\begin{eqnarray*}
    \cE_t := \bigcap_{i=1}^K \bigcap_{j\neq i}\bigcap_{d=1}^D   \left\{L^d_{i,j}(t, \delta) \leq \mu_i^d - \mu_j^d \leq U^d_{i,j}(t, \delta)\right\}, \ \text{ and } \ \cE = \bigcap_{t =1}^{\infty} \cE_{t},
\end{eqnarray*}
the lemma hereafter shows that with the choice of $\beta_i$'s in \eqref{eq:beta-single} and for $1$-subgaussian marginals, $\bP(\cE) \geq 1-\delta$. 
\begin{lemma}
\label{lem:correctness-single-beta}
   It holds that $\bP(\cE) \geq 1 - \delta.$
\end{lemma}
\begin{proof}
    Letting $$\tilde \cE_t := \bigcap_{i=1}^K\bigcap_{d=1}^D \left\{\lvert \muh_i^d(t) - \mu_i^d\lvert \leq \beta_i(t)\right\}, $$
    we have $\tilde \cE_t \subset \cE$. Indeed, on $\tilde \cE_t$, for any $i,j\in \bA$ and $d\leq D$, $$ \muh_i^d(t) - \muh_j^d(t) - \beta_i(t) - \beta_j(t) \leq \mu_i^d - \mu_j^d \leq \muh_i^d(t) - \muh_j^d(t) + \beta_i(t) + \beta_j(t), $$
which combined with $\beta_{i,j}(t) = \beta_i(t) + \beta_j(t)$ yields  $\tilde \cE_t \subset \cE_t$ so 
\begin{equation}
 \bP(\cE^\complement
) \leq \sum_{t=1}^\infty \bP(\cE_t^\complement) \leq \sum_{t=1}^\infty \bP(\tilde \cE_t^\complement).  
\end{equation}
Applying Hoeffding's inequality to the $1$-subgaussian marginals yields 
\begin{eqnarray*}
    \bP(\tilde \cE_t^\complement) &\leq& \sum_{i=1}^K \sum_{d=1}^D  \bP\left(\lvert\muh_i^d(t) - \mu_i^d\lvert > \beta_i(t)\right), \\
    &\leq& \sum_{i=1}^K \sum_{d=1}^D \sum_{s=1}^t \bP(\lvert\muh^d_{i, s} - \mu_i^d\lvert > \beta^{t, s}) \quad \text{ where } \beta^{t, s} = \sqrt{\frac{2}{s}\log\left(\frac{5KDt^4}{2\delta}\right)},\\
    \label{eq:ub-proba-cE-tilde}
    &\leq& \sum_{i=1}^K \sum_{d=1}^D \sum_{s=1}^t \frac{4\delta}{5KDt^4}, \\ &=& \frac{4\delta}{5t^3}.
\end{eqnarray*}
Finally, 
\begin{eqnarray*}
 \bP(\cE^\complement
) &\leq&  \sum_{t=1}^\infty \bP(\tilde \cE_t^\complement), \\
&\leq& \frac{4\delta}{5} \sum_{t=1}^\infty \frac{1}{t^3},\\
&\leq& \delta. 
\end{eqnarray*}
\end{proof}
We can now state the main theorem of this section. 
\begin{theorem}
Let $\eps_1 \geq 0, k\leq K$ and $\nu$ a bandit with $1$-subgaussian marginals. 
\algname run with the $\beta_i'$s of \eqref{eq:beta-single} and associated to the stopping time $\tau_{\eps_1}^k$ outputs a valid set and its expected sample complexity is upper-bounded as follows :  
$$ \bE_{\nu}(\tau_{\eps_1}^k) \leq {64\sqrt e H}\log\left( \frac{5KD}{2\delta}\right) + 
{256\sqrt e H}\log\left({256H}\right) + \frac{8\pi^2}{15} + 1,$$
with $H:= \sum_{a=1}^K{\max(\Delta_{a_t}, \eps_1, \omega_k)^{-2}}.$
\end{theorem}
\begin{proof}[Proof of \autoref{thm:sampcompl}]
The correctness follows from \autoref{lem:correctness-generic} combined with \autoref{lem:correctness-single-beta}. It remains to upper-bound $\bE(\tau_{\eps_1}^k)$. Note that this proof technique has been already used in \cite{kalyanakrishnan_pac_2012, kaufmann_information_2013, jun_anytime_2016}  for LUCB-like algorithms.
Let $n\geq 1$ to be specified later and 
\begin{eqnarray}
\cE(n) &=& \bigcap_{t\in [\frac12 n, n]} \cE_t.
\end{eqnarray}
Remark that 
\begin{eqnarray}
\label{eq:to_bound}
    \cE(n) \cap \{ \tau_{\eps_1}^k > n\} \tm{ holds } &\impl& 
    \sum_{t=1}^n \ind\{\{\tau_{\eps_1}^k> t\}\cap \cE(n)\} = n .
\end{eqnarray}
We will show that for some choice of $n$, the RHS of \eqref{eq:to_bound} will be strictly less than $n$ so the LHS does not hold. We proceed by upper-bounding the RHS
\begin{eqnarray}
\label{eq:exp_sampl_eq1}
\sum_{t=1}^n \ind\{\{\tau_{\eps_1}^k > t\}\cap \cE(n)\} &\leq& \frac n2 +   \sum_{t=\frac n2}^n \ind\{\{\tau_{\eps_1}^k > t\}\cap \cE(n)\}, \\&\leq& 
\frac n2 + \sum_{t=\frac n2}^n \ind\{ \{\tau_{\eps_1}^k > t\}\land \cE_t \}. 
\end{eqnarray}
From \autoref{lem:samplold}, \autoref{lem:sampl-complex-eps} and \autoref{lem:samplwm}, we have that for any $t\in [\frac n2 , n]$, 
$$ \{\tau_{\eps_1}^k > t\}\cap \cE_t \impl \max(\Delta_{a_t}, \eps_1, \omega_k) \leq 2\beta_{a_t, a_t}(t), $$ with $\beta_{a_t, a_t}(t) = 2\beta_{a_t}(t)$. Therefore, 
using this result back in \eqref{eq:exp_sampl_eq1} and letting $c_\delta := ({5KD}/{(2\delta)})^{1/4}$, $\tilde \Delta_a := \max(\Delta_a, \eps_1, \omega_k)$ yields 
\begin{eqnarray*}
  \sum_{t=1}^n \ind\{\{\tau_{\eps_1}^k > t\}\cap \cE(n)\} 
  &\leq& \frac n2  + \sum_{t=\frac n2 }^n\ind\{ \tilde \Delta_{a_t} \leq 4\beta_{a_t}(t) \}\\&\leq& \frac n2  + \sum_{t=\frac n2 }^n\sum_{a=1}^K\ind\{ (a_t = a) \land \tilde \Delta_a \leq 4\beta_{a}(t) \}, \\
  &\leq& \frac n2 + \sum_{a=1}^K\sum_{t=\frac12 n}^n \ind\{\{a_t = a\} \land \{T_{a}(t) \leq \frac{128}{\tilde \Delta_a^2}\log(c_\delta t)\}\}\\
  &\leq& \frac n2 + \sum_{a=1}^K \sum_{t=\frac n2}^n \ind\{\{a_t = a\} \land \{T_{a}(t) \leq \frac{128}{\tilde \Delta_a^2}\log(c_\delta n)\}\} \\
  &\leq& \frac n2  + \sum_{a=1}^K \frac{128}{\tilde \Delta_a^2}\log(c_\delta n)\\
  &\leq& \frac n2  + 128 H \log(c_\delta n), 
\end{eqnarray*}
where $H:= \sum_{a}\tilde \Delta_a^{-2}$. Then, choosing  $n$  such that 
$$\frac n2 + 128 H \log(c_\delta n) < n, $$
that is  
\begin{equation}
 n > T^\star := \inf \left\{ s\in \bN^\star : \frac{128 H \log(c_\delta s)}{s}< \frac12 \right\},   
\end{equation}
would yield 
\begin{equation}
    \sum_{t=1}^{n} \ind\{\{\tau_{\eps_1}^k > t\}\cap \cE(n)\} < n, 
\end{equation}
so $$\cE(n) \cap \{ \tau_{\eps_1}^k > n\} = \emptyset,$$ which means $$\{\tau_{\eps_1}^k > n\} \subset \cE(n)^\complement.$$
Therefore, for any $n>T^\star$, 
\begin{equation}
    \{\tau_{\eps_1}^k > n\} \subset \cE(n)^\complement.
\end{equation}
Thus, 
\begin{eqnarray*}
 \bE_{\nu}(\tau_{\eps_1}^k)  &=&    \bE_{\nu}\left(\tau_{\eps_1}^k \ind\{\tau_{\eps_1}^k \leq T^\star\} + \tau_{\eps_1}^k \ind\{\tau_{\eps_1}^k > T^\star \} \right)\\
 &\leq& T^\star  + \bE_{\nu}(\tau_{\eps_1}^k \ind\{\tau_{\eps_1}^k> T^\star \}\\
 &\leq& T^\star  + \sum_{n=T^\star + 1}^\infty \bP_\nu(\tau_{\eps_1}^k > n)\\
 &\leq& T^\star  + \sum_{n=T^\star + 1}^\infty \bP_\nu(\cE(n)^\complement), 
\end{eqnarray*}
using \eqref{eq:ub-proba-cE-tilde} and union bound yields, 

\begin{eqnarray*}
 \bP(\cE(n)^\complement) &\leq&  \sum_{t=\frac n2}^n \frac{4\delta}{5t^3},\\&\leq& \frac{4\delta}{5} \frac{(\nicefrac12)n}{(\nicefrac12)^3n^3},  \\
 &=& \frac{16\delta}{5} \frac{1}{n^2}.
\end{eqnarray*}
Then, 
\begin{eqnarray*}
   \bE_{\nu}(\tau_{\eps_1}^k)  &\leq&     T^\star  + \frac{16\delta}{5} \frac{\pi^2}{6}\\
   &\leq& T^\star +  \frac{8\pi^2}{15}.
\end{eqnarray*}
Upper-bounding $T^\star$ will conclude the proof. 
\begin{lemma} 
\label{lem:bound_T_star}
It holds that
    $$T^\star -1 \leq  \frac{1}{c_\delta} \exp\left(-W_{-1}\left( -\frac{1}{256c_\delta H} \right)\right) \leq {256\sqrt e H}\log\left({256c_\delta H }\right).$$
\end{lemma}
Finally, 
\begin{eqnarray}
 \bE_{\nu}(\tau_\delta) &\leq& {256\sqrt e H}\log\left({256(5KD/(2\delta))^{1/4} H}\right)  + \frac{8\pi^2}{15}  + 1,   \\
 &\leq&  
{64\sqrt e H}\log\left( \frac{5KD}{2\delta}\right) + 
{256\sqrt e H}\log\left({256H}\right) + \frac{8\pi^2}{15} + 1, 
\end{eqnarray}
which achieves the proof. 
\begin{remark}
The same technique could be applied to upper-bound $\bE_\nu(\tau_{\eps_1, \eps_2})$. 
\end{remark}
Now we prove \autoref{lem:bound_T_star}. 
    \begin{proof}[Proof of \autoref{lem:bound_T_star}]
        We have 
         \begin{eqnarray}
         \label{eq:eq_bound_T_star1}
             128 H\frac{\log(c_\delta s)}{s} < \frac12 &\equi& \frac{\log(c_\delta s)}{ s} < \frac{1}{256H} 
         \end{eqnarray}
         then, using \autoref{lem:ineq_log} yields 
         \begin{eqnarray*}
             \eqref{eq:eq_bound_T_star1} &\impl& \begin{cases}
                 s > 0 & \tm{ if } \frac{1}{256H} <c_\delta / e \\
                 0<s\leq \frac{1}{c_\delta} \; \tm{or}\; s \geq N^\star & \tm{ else, }
             \end{cases}
         \end{eqnarray*}
         with
         $$ N^\star = \frac{1}{c_\delta} \exp\left(-W_{-1}\left( -\frac{1}{256c_\delta H } \right)\right).$$
         Therefore, 
         $$ T^\star = \inf\left\{s \in \bN^\star : 128H\frac{\log(c_\delta s)}{s} < \frac12 \right\} \leq    \begin{cases}
            1  & \tm{ if }  \frac{1}{256H} <c_\delta / e\\
            \ceil{N^\star} & \tm{ else.}
         \end{cases} $$
         Using \autoref{lem:bound_W_1}, to upper bound $N^\star$ yields 
         \begin{equation}
             T^\star - 1 \leq {256\sqrt{e} H}\log_+\left( {256c_\delta H }\right),
         \end{equation}
         where $\log_+(x) = \max(0, \log(x))$. 
    \end{proof}
\end{proof}
\section{Technical Lemmas}
\label{sec:tech_lemmas}

\begin{lemma}
\label{lem:ineq_log}
    Let $a, b>0$. If $b < a/e$ then 
    $$ \frac{\log(ax)}{x} < b \impl  0<x \leq \frac1a \;\tm{or}\;x \geq \frac{1}{a}\exp\left(-W_{-1}\left(-\frac{b}{a}\right)\right).$$
Moreover, if $b\geq a/e$, then for any $x>0$, $\log(ax)/x \leq b$. 
\end{lemma}
\begin{proof}
    We have 
    \begin{eqnarray*}
        \frac{\log(ax)}{x} < b &\impl&  -\frac{1}{ax} \log\left(\frac{1}{ax}\right) < \frac ba \\&\impl& y \log(y) > - \frac{b}{a}, \quad y = \frac{1}{ax} \\
        &\impl& \frac{1}{ax} \geq  1 \; \tm{or} \; -\frac{b}{a}<y\log(y) < 0 
    \end{eqnarray*}
since $-b/a>-1/e$ and the negative branch $W_{-1}$ of the Lambert function is decreasing on $[-1/e, 0]$, 
\begin{eqnarray*}
 \frac{\log(ax)}{x} < b &\impl& 0<x \leq \frac{1}{a}  \; \tm{or} \; W_{-1}(y\log(y)) \leq W_{-1}(-b/a) \\
 &\impl& 0<x \leq \frac{1}{a}  \; \tm{or} \; \log(y) \leq W_{-1}(-b/a) \\
  &\impl& 0<x \leq \frac{1}{a}  \; \tm{or} \; ax \geq \exp(-W_{-1}(-b/a))\\
  &\impl& 0<x \leq \frac{1}{a}  \; \tm{or} \; x \geq \frac1a\exp(-W_{-1}(-b/a)). 
\end{eqnarray*}
Proving the second part of the lemma just follows from $\log(x) \leq x/e$. 
\end{proof}
The following lemma is taken from \cite{jourdan_top_2022}
\begin{lemma}[\cite{jourdan_top_2022}]
\label{lem:lambert_dom}
    For any $x \in [0, -e^{-1}]$, 
    \[  -\log(-x) + \log(-\log(-x)) \leq - W_{-1}(x) \leq -\log(-x) + \log(-\log(-x)) + \min\left\{ \frac12, \frac{1}{\sqrt{-x\log(-x)}}\right\} \]
\end{lemma}

\begin{corollary}
\label{lem:bound_W_1}
    Let $0<a<1/e$. It holds that 
    $$ \exp(-W_{-1}(-a)) \leq \frac{e^{1/2}}{a}\log\left(\frac{1}{a}\right).$$
\end{corollary}
We recall the following lemma which is taken from \cite{kaufmann_mixture_2021}. 
\begin{proof}
    Using \autoref{lem:lambert_dom} yields, 
    $$- W_{-1}(-a) \leq -\log(a) + \log(-\log(a)) + \frac12,$$
    and taking $\exp$ on both sides gives the result.
\end{proof}

\begin{lemma}
\label{lem:bound_part_tps}
    Let $\Delta^2>0$. Then, for $t\geq 2$, 
    $$ t \geq \frac{1}{\Delta^2}\log\left(2\log\left(\frac{3e^2}{2\Delta^2}\right)\right) \impl \frac{\log\log\left(e^4t\right)}{t} < \Delta^2.$$
\end{lemma}
   \begin{proof}
    We note that if $\Delta^2 \geq \frac{e}{3}$, then the result follows trivially since it can be easily checked that for $t\geq 2$,
    $$ \log\log(e^4t) \leq \frac{e}{3}t.$$
    Therefore, in the sequel, we assume $\Delta^2 < e/3$. 
        Let $$t_\Delta := \frac{1}{\Delta^2}\log\left(2\log\left(\frac{3e^2}{2\Delta^2}\right)\right),$$ and 
        $$ g(t) = t - \frac{1}{\Delta^2}{\log(\log(e^4t))}.$$
     Then, 
     $$g'(t) = 1 - \frac{1}{\Delta^2 t \log(e^4t)},$$
     and $g'(t) \geq 0$ for $t$ such that $\Delta^2 t\log(e^4t)\geq 1$. Using the Lambert function $W_0$, which is increasing on $[0, \infty)$,
        \begin{eqnarray}
             \Delta^2 t\log(e^4t) \geq 1 &\equi& e^4t\log(e^4t) \geq \frac{e^4}{\Delta^2} \\ &\equi& \log(e^4t) \geq W_0\left(\frac{e^4}{\Delta^2}\right)\\
             &\equi& t\geq t^0:=\frac1{e^4} \exp\left(W_0\left(\frac{e^4}{\Delta^2}\right)\right)
        \end{eqnarray}   
    and by definition of $W_0$, we have $$ W_0(x)\exp(W_0(x)) = x,$$ so 
    $$ \exp\left(W_0\left(\frac{e^4}{\Delta^2}\right)\right) = \frac{e^4}{\Delta^2} \frac{1}{W_0(e^4\Delta^{-2})}, $$
and therefore, 
$$
t^0 = \frac{1}{\Delta^2} \frac{1}{W_0(e^4\Delta^{-2})}.
$$
We will show that $t_\Delta > t^0$. Indeed, since $W_0$ is increasing, 
    \begin{eqnarray*}
        \frac{1}{\Delta^2} > 3/e &\impl& W_0\left(\frac{e^4}{\Delta^2}\right) \geq W_0(3e^3) = 3 \\&\impl& 
        \frac{1}{\Delta^2} \frac{1}{W_0(e^4\Delta^{-2})} \leq \frac{1}{3} \frac{1}{\Delta^2} ,
    \end{eqnarray*}
    that is 
    \begin{equation}
\label{eq:upper-bound-t0}
t^0 \leq  \frac13 \frac{1}{\Delta^2} . 
    \end{equation}
On the other side, 
\begin{eqnarray*}
    \frac{1}{\Delta^2} > 3/e &\impl& \log\left(2\log\left(\frac{3e^2}{2\Delta^2}\right)\right) > \log(2\log(9e/2)) \\
    &\impl& t_\Delta \geq \frac{1}{\Delta^2}\log(2\log(9e/2))> \frac13 \frac1{\Delta^2}
\end{eqnarray*}
Therefore, \begin{equation*}
   t^0 \leq \frac13\frac{1}{\Delta^2} <\frac{\log(2\log(9e/2))}{\Delta^2} \leq t_\Delta. 
\end{equation*}
 Thus, we have shown that $t^0 \leq t_\Delta$  and for any $t\geq t_\Delta$, $g'(t) \geq 0$  so 
 \begin{equation}
 \forall t \geq t_\Delta, \; g(t) \geq g(t_\Delta).    
 \end{equation}
 Showing that  $g(t_\Delta) > 0$, will conclude the proof. 
Letting $a = 3e^2/2$, we have 
 \begin{eqnarray}
 g(t_\Delta) > 0   &\equi& \frac{1}{\Delta^2}\log(2\log(a/\Delta^2)) - \frac{1}{\Delta^2}\log(\log(e^4t_\Delta)) > 0 \\
  &\equi& \log(2\log(a/\Delta^2)) - \log(\log(e^4t_\Delta)) > 0 \\
   &\equi& 2\log(a/\Delta^2) - \log(e^4t_\Delta) > 0 \\
   &\equi& \log(a/\Delta^2) - \log(e^4t_\Delta\Delta^2  /a) >0 \\
    &\equi& \frac{a}{\Delta^2} - \frac{e^4}{a}{\Delta^2} t_\Delta > 0 \\
    \label{eq:sign_g}
    &\equi& \frac{a}{\Delta^2} - \frac{e^4}{a}\log(2\log(a/\Delta^2))> 0,
\end{eqnarray}
then,  observing that for $x\geq 12$, 
$$ \log(2\log(x)) \leq \frac{x}{e^2},$$
and since $$ a/\Delta^{2}  > (3e^2/2) \times (3/e) > 12,$$
we have 
\begin{equation}
\label{eq:bound-log-log}
    \log(2\log(a/\Delta^2)) \leq \frac1{e^2} \frac{a}{\Delta^2}
\end{equation}
so, using \eqref{eq:bound-log-log} yields that the LHS of \eqref{eq:sign_g} is larger than 
$$\frac{3}{2}e^2 \frac1{\Delta^2} - e^2 \frac1{\Delta^2}, $$
which is always positive. 
Therefore, \begin{equation*}
    \forall t \geq t_\Delta,\; g(t_\Delta) > 0 , 
\end{equation*}
that is 
\begin{equation}
    \forall t \geq t_\Delta, \; \frac{\log\log(e^4t)}{t} < \Delta^2.
\end{equation}
    \end{proof}
\begin{lemma}
\label{lem:bound_inv_beta}
   Let $\delta \in (0, 1), \Delta> 0$ and $c>0$. Let $f : t \mapsto \sqrt{ \frac{g(\delta) + c\log\log(e^4t)}{t}}$
    where $g$ is a non-negative function. Then, for any $\alpha \in (0,1)$ and $t\geq 2$, 
    \begin{equation}
        t \geq \frac{1}{\Delta^2} \left( \frac{1}{\alpha} g(\delta) + \frac{c}{1-\alpha}\log_+\left(2\log\left(\frac{c}{(1-\alpha)\Delta^2}\right)\right)\right) \impl f(t) < \Delta. 
    \end{equation}
    \end{lemma}
        \begin{proof}
       Letting $t\geq 2$, we have 
        \begin{equation}
\label{eq:eq_bound_inv_beta_1}
            t \geq t_1:= \frac1\alpha\frac{1}{\Delta^2} g(\delta) \impl \frac{g(\delta)}{t} \leq \alpha \Delta^2. 
        \end{equation}
        Furthermore, using \autoref{lem:bound_part_tps} yields 
        \begin{equation}
\label{eq:eq_bound_inv_beta_2}
            t \geq t_2:= \frac{c}{(1- \alpha)\Delta^2} \log_+\left(2\log\left(\frac{3e^2}{2(1-\alpha)\Delta^2}\right)\right) \impl \frac{\log\log(e^4t)}{t} \leq (1- \alpha)\Delta^2/c.
        \end{equation}
        Combining \eqref{eq:eq_bound_inv_beta_1} and \eqref{eq:eq_bound_inv_beta_2} yields for $t\geq 2$,
        \begin{eqnarray*}
            t \geq \max(t_1, t_2) \impl f(t)^2 < \Delta^2, 
        \end{eqnarray*}
        so \begin{equation}
            t \geq t_1 + t_2\geq \max(t_1, t_2) \impl f(t) < \Delta. 
        \end{equation}
    \end{proof}

\begin{lemma}
\label{lem:bound_inv_beta_ek}
Let $\Delta > 0$ and $\delta \in (0, 1)$. Let 
 \[ f(t) := 4\sqrt{\frac{2 C^g(\log(1/\delta)/2) + 4\log\log(e^4t)}{t}}.\]
 Then 
$$ \inf\{n\geq 2 : f(n) < \Delta \} \leq \frac{88}{\Delta^2}\log\left(\frac{4}{\delta}\log\left(\frac{12e}{\Delta}\right)\right).$$
\end{lemma}
\begin{proof}
We have 
\begin{eqnarray*}
  f(t) = \sqrt{\frac{32 C^g(\log(1/\delta)/2) + 64\log\log(e^4t)}{t}}.
\end{eqnarray*}
Therefore, letting 
$$ g(\delta):= 32 C^g(\log(1/\delta)/2) \quad \text{ and } \quad  c= 64$$ and further using \autoref{lem:bound_inv_beta} yields for any $\alpha \in (0, 1)$ and $t\geq 2$, 
$$ t\geq t_{\alpha} \impl f(t) < {\Delta},$$
where 
$$ t_{\alpha} := \frac{1}{\Delta^2} \left(\frac{32}{\alpha}C^g(\log(1/\delta)/2) + \frac{64}{1-\alpha}\log\left(2\log(96e^2(1-\alpha)^{-1}\Delta^{-2})\right)\right).$$
   Since $C^g(x)\approx x + \log(x)$ \cite{kaufmann_mixture_2021},  and $\log(x) \leq x/e$ we have 
   $$ \Delta^2 t_{\alpha} \leq \left({16}  + \frac{16}{e}\right)\frac{1}{\alpha}\log(1/\delta) +  \frac{64}{1-\alpha}\log(2\log(96e^2(1-\alpha)^{-1}\Delta^{-2})).$$
Taking $\alpha = \alpha^\star$ such that $$\left({16}  + \frac{16}{e}\right)\frac{1}{\alpha^\star} = \frac{64}{1-\alpha^\star},$$
that is setting \begin{equation*}
    \alpha^\star = \frac{1+e}{1+5e}, 
\end{equation*}
yields
$$\Delta^2 t_{\alpha^\star} \leq \frac{64}{1 - \alpha^\star}\log\left( \frac2\delta \log\left( \frac{96e^2}{(1-\alpha^\star) \Delta^2}\right)\right).$$
By numerical evaluation, 
$$ \frac{64}{1 - \alpha^\star}  \approx 86 < 88 \quad \text{ and } \quad \frac{96}{1-\alpha^\star} \approx 130 <  12^2,$$
so 
\begin{equation}
\Delta^2 t_{\alpha^\star} < 88\log\left( \frac4\delta \log\left( \frac{12e}{\Delta}\right)\right). 
\end{equation}
Therefore, 
putting these results together, for $t\geq 2$
$$t\geq  t_{\star} := \frac{88}{\Delta^2}\log\left( \frac4\delta \log\left( \frac{12e}{\Delta}\right)\right) \impl f(t) < \Delta$$ which yields
\begin{equation}
\inf\{n\geq 2: f(n) < \Delta \} \leq \max(2, t_\star).    
\end{equation}
\end{proof}

\section{Implementation and Additional Experiments}
\label{sec:impl_exp}
In this section, we give additional details about the experiments and additional experimental results.
\subsection{Implementation}
\label{subsec:implementation}
\paragraph{Setup} We have implemented the algorithms mainly in \texttt{C++17} compiled with GCC12 and interfaced with \texttt{python} through the \texttt{cython} package. The experiments are run on an ARM64 8GB RAM/8 core/256GB disk storage computer. 
For the function $C^g$ we have used the approximation $C^g(x) \approx x + \log(x)$  which is usually admitted \cite{kaufmann_mixture_2021}.  
For the experiments on real-world scenario we generate a certain number of seeds (usually 2000) and we use  a different seed for each run on the same bandit.  This procedure is identical for every experiment where we report the average sample complexity on the same bandit. To assess the robustness of our algorithm, the experiments on the synthetic dataset consisted in randomly uniformly sampling some bandit means for each configuration. For each sampled bandit, the algorithms compared are run once on the same instance and we note their empirical sample complexity. Finally, we report the average sample complexity across all the bandits of the same configuration.  

\paragraph{Time and memory complexity} The time complexity of $\varepsilon_1$-APE-$k$ is $\mathcal{O}(K^2D)$ and its memory complexity is $\mathcal{O}(K^2)$. The main computational bottleneck is the computation of the $M(i,j,t)$ for each $(i,j) \in K \times K$, which requires a triple-nested for-loop over $[K]\times [K]\times[D]$.	
	To give an idea of the runtime, a single run on a random Bernoulli instance with $K=1000, D=10$ takes around 4 minutes for $0.1-$APE-$1000$ on a personal computer (a single 3GHz ARM core used, 8 GB RAM, 256 GB disk storage) with $\delta=0.01$ and no particular optimization. Due its fully sequential nature, our algorithm may have a higher computational cost compared to uniform sampling strategies. However, in our implementation the most time-consuming operation was actually collecting a sample from the selected arm(s), especially for multivariate Gaussians. So that finally, in the experiments, our algorithm which ultimately require less samples had in practice a similar computational cost compared to \algauername which uses uniform sampling. 
\paragraph{Adaptation to bandits with marginals of different scaling} We have presented the algorithm and the results specialized to the case where all the marginals are $1$-subgaussian. Indeed our results can be simply extended where the marginals are instead all $\sigma$-subgaussian. Furthermore, there is a simple way to adapt the algorithm to the case where the marginals have different \emph{known} subgaussianity parameter (i.e different scaling) but they are the same for every arm. The idea is to rescale each observation with the subgaussianity parameters. Let $\boldsymbol{\sigma}:= \sigma_{1}, \dots \sigma_{D}$, $\sigma_i>0$. Assuming that the marginal distributions of each arm are respectively $\sigma_{1}, \dots \sigma_{D}$-subgaussian, each observation $\bX_{A_t}, s$ from arm $A_t$ will be rescaled component-wise to $X_{A_t,s}^d / \sigma_{d}$ before being given to the algorithm. It is easy to see that this rescaling does not change the Pareto set since all the means are divided by the same values coordinate-wise. 
Furthermore, by defining 
\begin{eqnarray*}
    \M^{\boldsymbol{\sigma}}(i,j) &:=& \max_d \left(\frac{\mu_i^d - \mu_j^d}{\sigma_{d}}\right), 
\end{eqnarray*}

and  $\m^{\boldsymbol{\sigma}}(i,j) =: - \M^{\boldsymbol{\sigma}}(i,j)$, all the results proved for $1$-subgaussian distributions still holds using $\m^{\boldsymbol{\sigma}}$ and $\M^{\boldsymbol{\sigma}}$ in the definition of the gaps (Section 4). 
\subsection{Data processing}
\label{subsec:data_processing}

\paragraph{Dataset} The dataset is extracted from \cite{munro_safety_2021} and some processing steps are applied to compute the covariance matrix of the distribution. First, as observed in \cite{munro_safety_2021}, the 3 immunogenicity indicators extracted are weakly correlated, therefore, we assume the covariance matrix to be diagonal. To compute the variance of the marginals, we use the log-normal assumption as assumed for the data reported in \cite{munro_safety_2021}. 
Using this log-normal assumption 
the authors have provided for each arm and each indicator: the geometrical mean, the sample size and a $95\%$ confidence interval on the geometrical mean based on the {central limit theorem}. 

 For each of the $K=20$ arms (combination of three doses), we use these information to compute the sample variance of each immunogenicity indicator.
Moreover, we compute the arithmetic average of the $\log$ outcomes which is obtained by taking the $\log$ (base $e$) of the geometrical empirical mean:
\begin{eqnarray*}
 \bar x &=& \log(\bar x_{\text{geometrical}}),\\
&=& \log\left( \left(\prod_{i=1}^n x_i\right)^{1/n}\right),\\
&=& n^{-1}\sum_{i=1}^n \log(x_i), 
\end{eqnarray*}
where $x_1, \dots, x_n$ are the observations which are assumed to be log-normal.
 $\bar x$ represents by assumption the empirical mean of a Gaussian distribution, which we use as a proxy for its true, unknown mean. From this, we built a bandit model where each arm is a 3-dimensional Gaussian distribution with independent coordinates, whose means are given by the corresponding mean estimates (reported in \autoref{tab:arith_mean}) and in which the variance of each indicator is the pooled variance over the different arms (given in \autoref{tab:pooled_var}). 
 Sampling an arm in this bandit simulates the measurement of the  ($\log$ of the) 3 immunogenicity criteria in consideration on a new patient.
 
 The 20 arms are classified into two groups. Each three/four-letters acronym denotes a vaccine candidate. Prime BNT/BNT corresponds to giving BNT as first and second dose and similarly for Prime ChAd/ChAd. For example ChAd in the group Prime BNT/BNT means to give BNT as first and second dose and ChAd as third dose (booster).

\begin{table}[H]
\caption{Table of the empirical arithmetic mean of the log-transformed immune response for three immunogenicity indicators. Each acronym corresponds to a vaccine. There are two groups of arms corresponding to the first 2 doses: one with prime BNT/BNT (BNT as first and second dose) and the second with prime ChAd/ChAd (ChAd as first and second dose). Each row in the table gives the values of the 3 immune responses for an arm (i.e. a combination of three doses).}
\begin{center}
\begin{tabular}{|c|c|c|c|c|}
\hline 
\hline 


\multirow{2}{*}{Dose 1/Dose 2} & \multirow{2}{*}{Dose 3 (booster)}  &\multicolumn{3}{c|}{Immune response}\\

 &  & Anti-spike IgG &  $\text{NT}_{50}$ &  cellular response\\\hline 
    \multirow{10}{*}{Prime BNT/BNT}&ChAd     &                      9.50 &                                       6.86 &                                 4.56 \\
&NVX      &                      9.29 &                                       6.64 &                                 4.04 \\
&NVX Half &                      9.05 &                                       6.41 &                                 3.56 \\
&BNT      &                     10.21 &                                       7.49 &                                 4.43 \\
&BNT Half &                     10.05 &                                       7.20 &                                 4.36 \\
&VLA      &                      8.34 &                                       5.67 &                                 3.51 \\
&VLA Half &                      8.22 &                                       5.46 &                                 3.64 \\
&Ad26     &                      9.75 &                                       7.27 &                                 4.71 \\
&m1273    &                     10.43 &                                       7.61 &                                 4.72 \\
&CVn      &                      8.94 &                                       6.19 &                                 3.84 \\
\hline
 \multirow{10}{*}{Prime ChAd/ChAd}&ChAd     &                      7.81 &                                       5.26 &                                 3.97 \\
&NVX      &                      8.85 &                                       6.59 &                                 4.73 \\
&NVX Half &                      8.44 &                                       6.15 &                                 4.59 \\
&BNT      &                      9.93 &                                       7.39 &                                 4.75 \\
&BNT Half &                      8.71 &                                       7.20 &                                 4.91 \\
&VLA      &                      7.51 &                                       5.31 &                                 3.96 \\
&VLA Half &                      7.27 &                                       4.99 &                                 4.02 \\
&Ad26     &                      8.62 &                                       6.33 &                                 4.66 \\
&m1273    &                     10.35 &                                       7.77 &                                 5.00 \\
&CVn      &                      8.29 &                                       5.92 &                                 3.87 \\
\hline
\end{tabular}
\end{center}
\label{tab:arith_mean}
\end{table}
\begin{table}[ht]
\caption{Pooled variance of each group.}
\begin{center}
\begin{tabular}{|c|c|c|c|}
\hline 
\hline 
\multirow{2}{*}{} & \multicolumn{3}{c|}{Immune response}\\
 & Anti-spike IgG &  $\text{NT}_{50}$ &  cellular response\\\hline 
    Pooled sample variance&  0.70&0.83& 1.54\\\hline 
\end{tabular}
\end{center}
\label{tab:pooled_var}
\end{table}
\subsection{Additional experiments}
\label{subsec:additional_experiments}
\subsubsection{Additional experiments for $\eps_1$-APE-$k$}
\begin{figure}[t]
     \centering
     \begin{subfigure}[b]{0.49\textwidth}
         \centering
    \includegraphics[width=\textwidth]{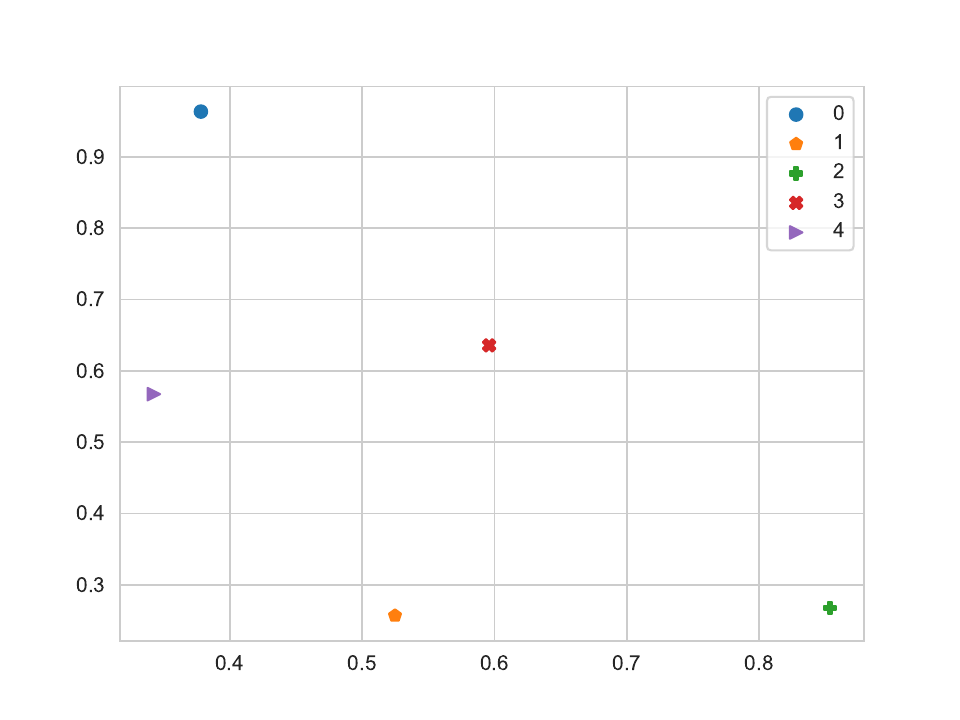}
         \caption{Instance}
         \label{fig:config-avantage-APE}
     \end{subfigure}
     \hfill
     \begin{subfigure}[b]{0.49\textwidth}
         \centering
    \includegraphics[width=\textwidth]{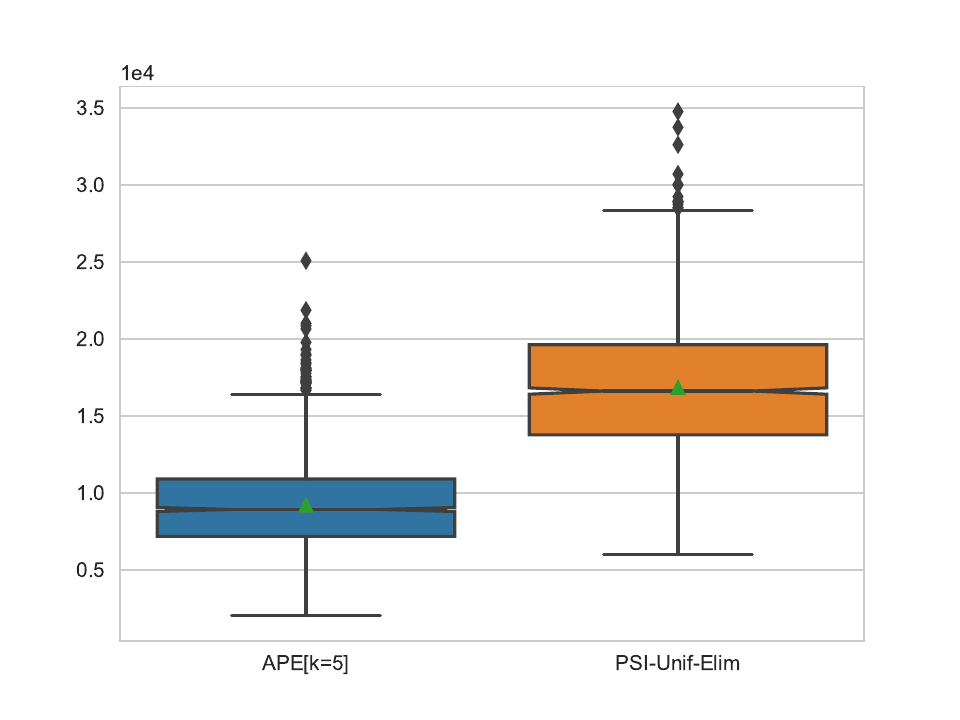}
         \caption{Sample complexity}
         \label{fig:appx-new-instance-ape-k-sc}
     \end{subfigure}
        \caption{For this instance (left) $\cS^\star = \{0,2, 3\}$ and on the right is the average sample complexity over the trials on the same instance.}
        \label{fig:appx-new-instance-ape-k-results}
\end{figure}

In this section we show that for some instances our algorithm can require up to 3 times less samples compared to \algauername. This is due to the strategy of \algauername which continue sampling arms identified as optimal until there are shown not to dominate any arm in the active set. For example on \autoref{fig:config-avantage-APE}, the optimal arm 2 is "easy" to identify as such. However, since it slightly dominates the sub-optimal arm 1, \algauername should continue sampling arm 2 until arm 1 is removed from the active set ( likely this will happen when the algorithm "sees" that arm 1 is dominated by arm 3).  We would expect our adaptive sampling rule to avoid this behaviour. 

\autoref{fig:appx-new-instance-ape-k-sc} shows that APE takes nearly half the average sample complexity of \algauername on this instance. In particular, \autoref{tab:average-pull-ratio} shows the average number of pulls taken  by \algauername divided by the average number of pulls taken by $0$-APE-$K$ for each arm. We can observe that the major difference in sample complexity is due to arm 2 being pulled nearly 6 times more by \algauername than \algname.  

\begin{figure}[H]
    \centering
    \includegraphics[width=0.6\linewidth]{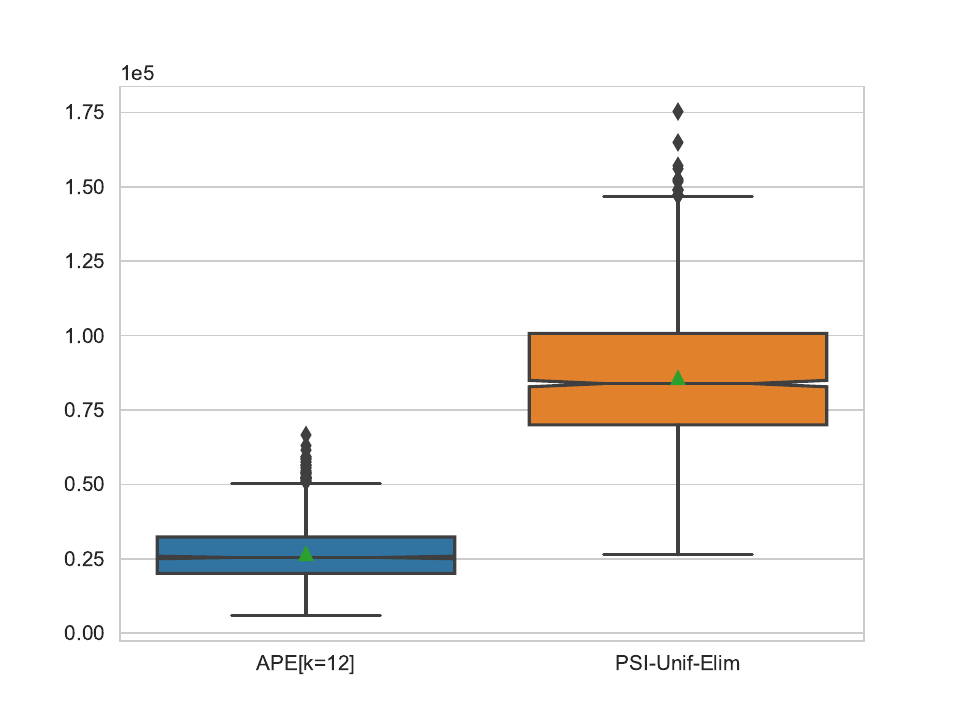}
    \caption{Average sample complexity on a specific instance with $K=12, D=10$ and $|\cS^\star|=11$}
    \label{fig:appx-new-instance-higher-dim}
\end{figure}

\begin{table}[t]
    \centering
    \begin{tabular}{|c|c|c|c|c|c|}
    \hline 
      Arm    &  0 & 1 & 2 & 3 & 4 \\\hline 
      Average ratio of pulls   & 3.08 & 1.36 & 5.68 & 1.28 & 1.40\\
      \hline
    \end{tabular}
    \caption{Average number of pulls taken  by \algauername divided by the average number of pulls taken by $0$-APE-$K$ for each arm.}
    \label{tab:average-pull-ratio}
\end{table}
By increasing the number of arms and the dimension we can generate instances similar to \autoref{fig:config-avantage-APE} where the gap between our algorithm and \algauername is even larger. We chose a specific instance where $K=12$, $D=10$ and there are 11 optimal arms. 
On this instance (\autoref{fig:appx-new-instance-higher-dim}), we can see that our algorithm uses 3 times less samples than \algauername. 

Finally, combining these additional experiments with the results of \autoref{tab:bayes-commun} we observe that on average $\eps_1$-APE-$k$ performs nearly $20\%$ better than \algname but there are some instances where the gap can be even larger. Of course, this also means that there should exist instances in which the improvement is smaller than $20\%$ to compensate for instances like \autoref{fig:config-avantage-APE}. But we note that instances like \autoref{fig:config-avantage-APE} are very unlikely to be generated randomly so they should only be a few in the 2000 instances used for \autoref{fig:config-avantage-APE}. So that $20\%$ is fairly representative of the average improvement on ``normal instances" (excluding instances like \autoref{fig:config-avantage-APE} where the improvement can be way larger).  

\subsubsection{$(\eps_1, \eps_2)$-APE}

\begin{figure}[t]
     \centering
     \begin{subfigure}[b]{0.49\textwidth}
         \centering
    \includegraphics[width=\textwidth]{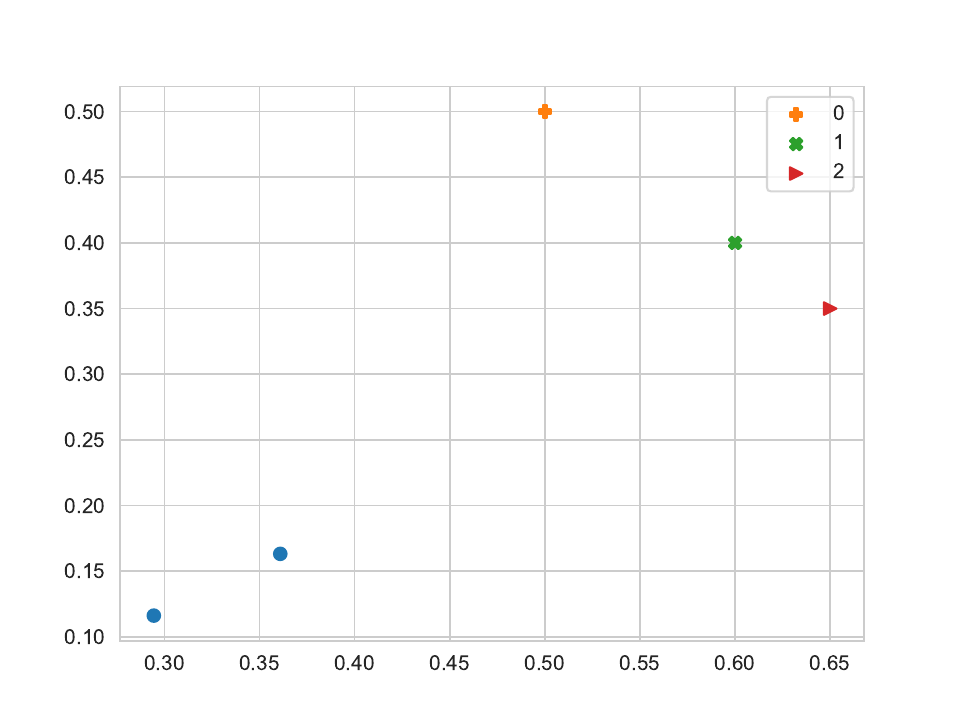}
         \caption{Instance}
         \label{fig:config_eps_2}
     \end{subfigure}
     \hfill
     \begin{subfigure}[b]{0.49\textwidth}
         \centering
    \includegraphics[width=\textwidth]{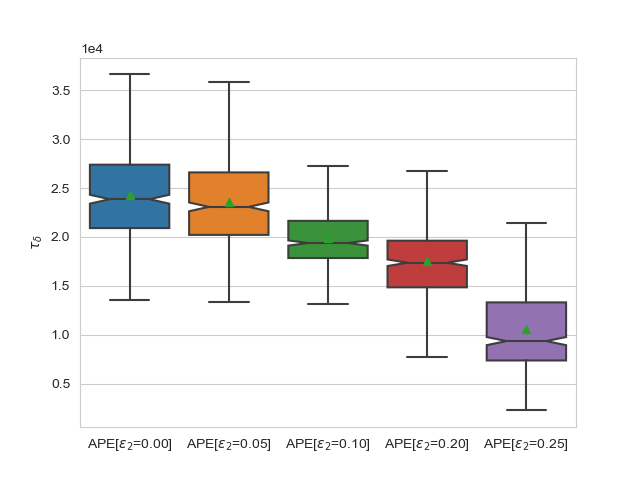}
         \caption{Sample complexity}
         \label{fig:appx-C-results-result}
     \end{subfigure}
        \caption{For this instance (left) $\cS^\star = \{0,1,2\}$. The difference in on \texttt{x} and \texttt{y} axis is $0.1$ between arm 0 and 1 and $0.05$ between arm 1 and 2. The rightmost figure is the sample complexity averaged over 2000 trials.}
        \label{fig:appx-C-results}
\end{figure}

We investigate the empirical behavior of $({\eps_1, \eps_2})$-APE for identifying an $(\eps_1, \eps_2)$-cover.  We set $\eps_1=0, \delta=0.01$ and we test different values of $\eps_2 \in \{0, 0.05, 0.1, 0.2, 0.25\}$. We average the results over 2000 independent trials with different seeds on the same instance (\autoref{fig:config_eps_2}). We use multi-variate Bernoulli with independent marginals. The instance of \autoref{fig:config_eps_2} is a toy example where $(\eps_1, \eps_2)$-covering can be meaningful and reduce the sample complexity. The 3 {Pareto optimal} vectors are chosen by hand and the last 2 vectors are randomly uniformly generated. 

\begin{figure}[htb]
     \centering
     \begin{subfigure}[b]{0.49\textwidth}
         \centering
    \includegraphics[width=\textwidth]{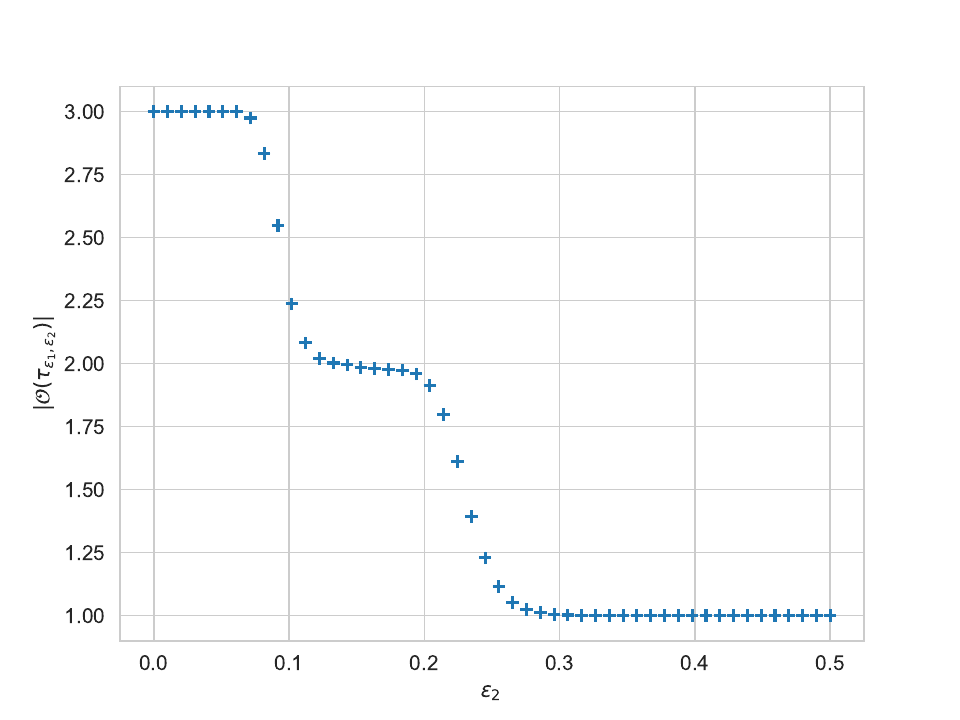}
         \caption{Average $\lvert\cO({\tau_{\eps_1, \eps_2}})\lvert$  vs $\eps_2$}
         \label{fig:appx-C-average_card_O}
     \end{subfigure}
     \hfill
     \begin{subfigure}[b]{0.49\textwidth}
         \centering
    \includegraphics[width=\textwidth]{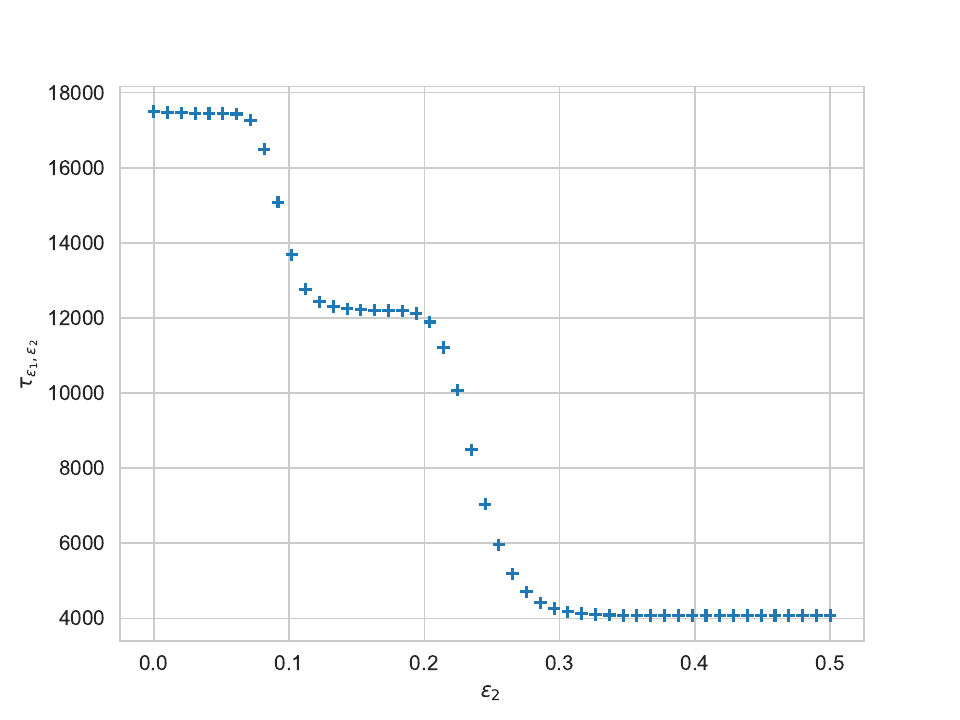}
         \caption{Sample complexity vs $\eps_2$}
         \label{fig:appx-C-sample_complexity_vs_eps_2}
     \end{subfigure}
        \caption{On the left is the average length of $\cO({\tau_{\eps_1, \eps_2}})$ (over the different runs) versus $\eps_2$ and on the right is the empirical sample complexity (averaged over the runs) versus $\eps_2$. The empirical probability of error was equal to zero.}
        \label{fig:appx-C-results_eps_2}
\end{figure}

\begin{figure}[htb]
     \centering
     \begin{subfigure}[b]{0.32\textwidth}
         \centering
    \includegraphics[width=\textwidth]{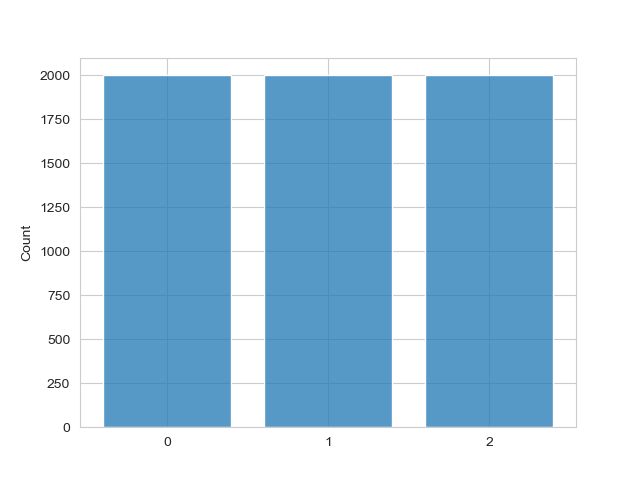}
         \caption{Average $\eps_2 = 0.05$}
         \label{fig:appx-C-freq-005}
     \end{subfigure}
     \hfill
     \begin{subfigure}[b]{0.32\textwidth}
         \centering
    \includegraphics[width=\textwidth]{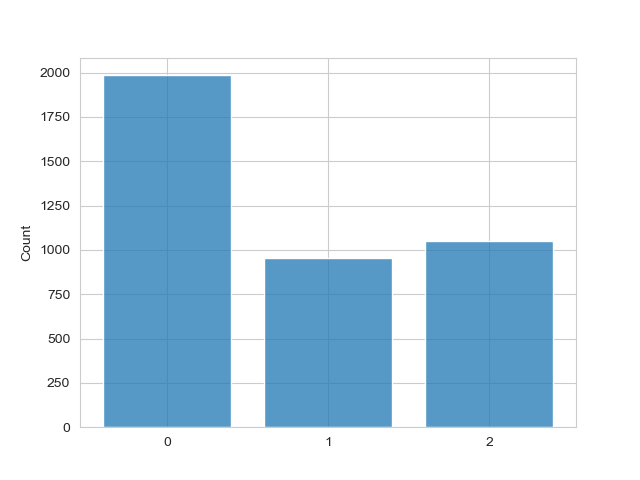}
         \caption{$\eps_2 = 0.15$}
         \label{fig:appx-C-freq-015}
     \end{subfigure}
     \hfill
        \begin{subfigure}[b]{0.32\textwidth}
         \centering
    \includegraphics[width=\textwidth]{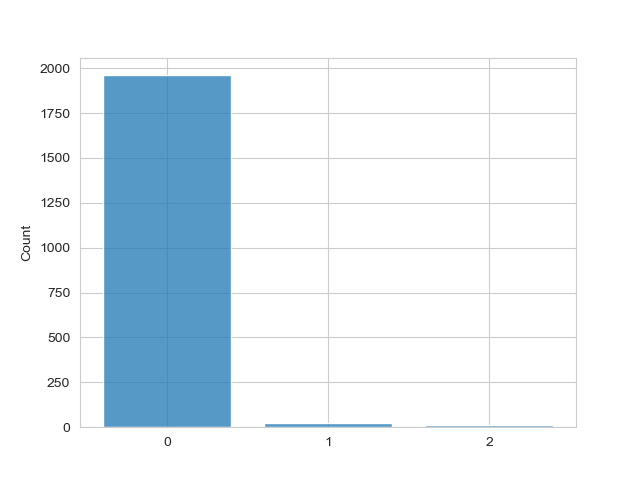}
         \caption{ $\eps_2 = 0.4$}
         \label{fig:appx-C-freq-04}
     \end{subfigure}
        \caption{Histogram of the number of times each arm is in the recommended set over the 2000 runs for 3 values of $\eps_2$. }
        \label{fig:appx-C-freq}
\end{figure}

We can observe on \autoref{fig:appx-C-results-result} that the sample complexity decreases as $\eps_2$ increases. 
This is further confirmed in \autoref{fig:appx-C-results_eps_2} which shows the empirical sample complexity and the average size of the recommended cover versus $\eps_2$ for 50 equally-spaced values of $\eps_2$ between $0$ and $1/2$.  The drops observed in \autoref{fig:appx-C-sample_complexity_vs_eps_2} correspond to the values of $\eps_2$ for which \algname removes an {optimal} arm from the cover to save some samples (\autoref{fig:appx-C-average_card_O}). A major decrease in the sample complexity corresponds more or less to an arm being removed from the recommended set. We observe in \autoref{fig:appx-C-freq} the histogram of occurrence of each arm in the recommended set for 3 values of $\eps_2$ corresponding more or less to  the middle of each plateau. We can see that for $\eps_2 = 0.15$, arm 0 is always recommended, but the others are recommended on half of the runs. For $\eps_2 =0.4$, the algorithm nearly always recommend arm 0, which as the largest $\omega_i$ term (i.e the easiest to identify as optimal).



The plateau in the sample complexity for large values of $\eps_2$ ($>0.3$) is explained by the fact the algorithm needs to identify at least one {optimal} arm (which is reflected in the size of the returned set \autoref{fig:appx-C-average_card_O}).
Indeed, for $\eps_1=0$ fixed, an algorithm for $(\eps_1, \eps_2)$-covering still need to assert that the arms in the recommended set are truly optimal which will require some samples even when $\eps_2$ is very large. Thus, for $\eps_2 >0.3$ the algorithm need to identify at least one optimal arm and we can see on \autoref{fig:appx-C-average_card_O} that for these values of $\eps_2$, the recommended set contains only one optimal arm. Actually we can observe empirically that the "limit" sample complexity observed in \autoref{fig:appx-C-sample_complexity_vs_eps_2} is close to the average sample complexity of $0$-APE-$1$ on the same instance ($4073$ samples). 

\subsubsection{Comparing $\eps_1$-APE-$k$ to an adaptation of \algauername}
In this section, we compare $\eps_1$-APE-$k$ to an adaptation of \algauername which stops earlier if at least $k$ optimal arms have been identified. The pseudo-code of the algorithm is given in \autoref{alg:heuristic-auer}. As shown in \cite{auer_pareto_2016}, arms in $P_1(t)$ are already identified as optimal but when the goal is to identify the Pareto set, some of them (namely $P_1(t)\backslash P_2(t)$) need to be sampled again until all the arms they potentially dominate are removed from $A(t)$ and only then those arms will belong to $P_2(t)$ and will be removed from the active set. However, for the $k$-relaxation, identifying $k$ optimal arms is enough so the algorithm can stop as soon as $\lvert P(t-1) \cup P_1(t)\lvert \geq k$. It this never occurs, then the algorithm will follow the initial stopping condition, that is to stop when $A(t) = \emptyset$. 
\begin{algorithm}[htb]
\caption{Adaptation of \algauername for the $\eps_1$-APE-$k$ objective}\label{alg:heuristic-auer}
\KwData{parameter $\eps_1\geq 0, k\leq K$}
\Input{$A(0) = \bA$, $t \gets 1$, $P(0) = P_1(0) = P_2(0) = \emptyset$ } 
 \For{$t=1,2,\dots$}{
 \Sample each arm in $A(t-1)$ once \;
 $A(t) \gets  \left\{i \in A(t-1): \forall \; j\in A(t-1), \mh(i,j,t)\leq \beta_{i,j}(t)\right\} $\; 
$P_1(t) \gets  \left\{ i \in A(t): \forall\; j\in A(t)\setminus \{i\}, \Mh(i,j,t) +\eps_1 \geq  \beta_{i,j}(t) \right\}$\;
$P_2(t) \gets \left\{ j \in P_1(t): \nexists \;i \in A(t)\setminus P_1(t) \tm{ s.t }  \Mh(i,j,t) +\eps_1\leq \beta_{i,j}(t)\right\}$\;
$A(t)\gets A(t) \setminus P_2(t)$ and $P(t) \gets  P(t-1) \cup P_2(t)$\;
\If{$ A(t) = \emptyset$}{
\textbf{break} and output $P(t)$\;
}
\If{$\lvert P(t-1) \cup P_1(t)\lvert \geq k$}{
\textbf{break} and output $P(t-1) \cup P_1(t)$\;
}
 }
\end{algorithm}

\begin{figure}[H]
     \centering
     \begin{subfigure}[b]{0.49\textwidth}
         \centering
    \includegraphics[width=\textwidth]{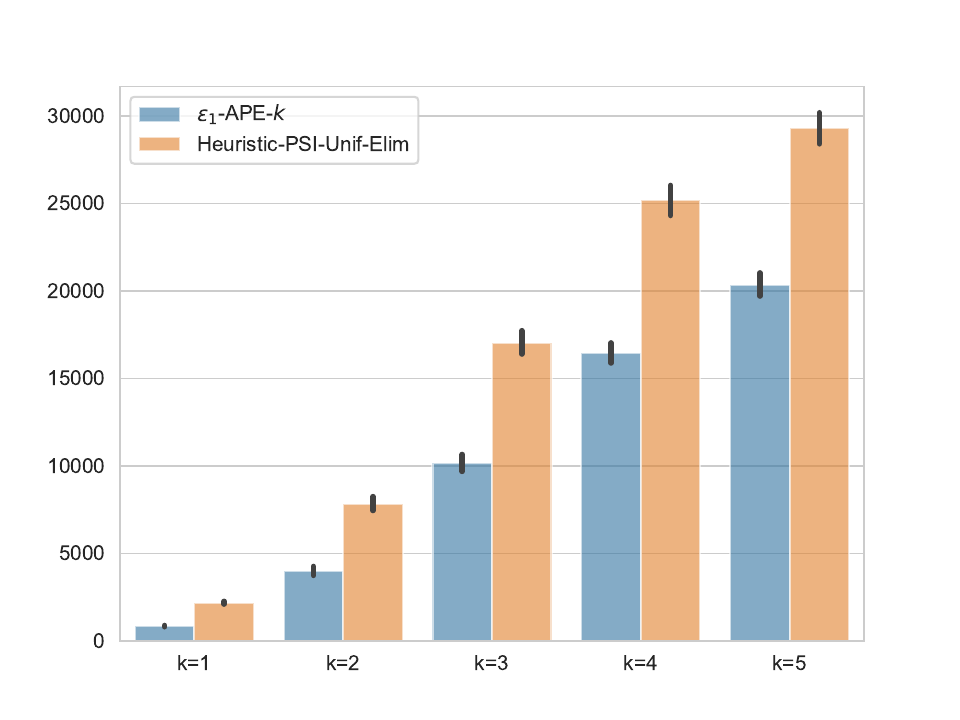}
         \caption{$K=10, D=2, \eps_1= 0.05$}
         \label{fig:appx-comp-ape-heuristic-K5k5-eps-001}
     \end{subfigure}
     \hfill
     \begin{subfigure}[b]{0.49\textwidth}
         \centering
    \includegraphics[width=\textwidth]{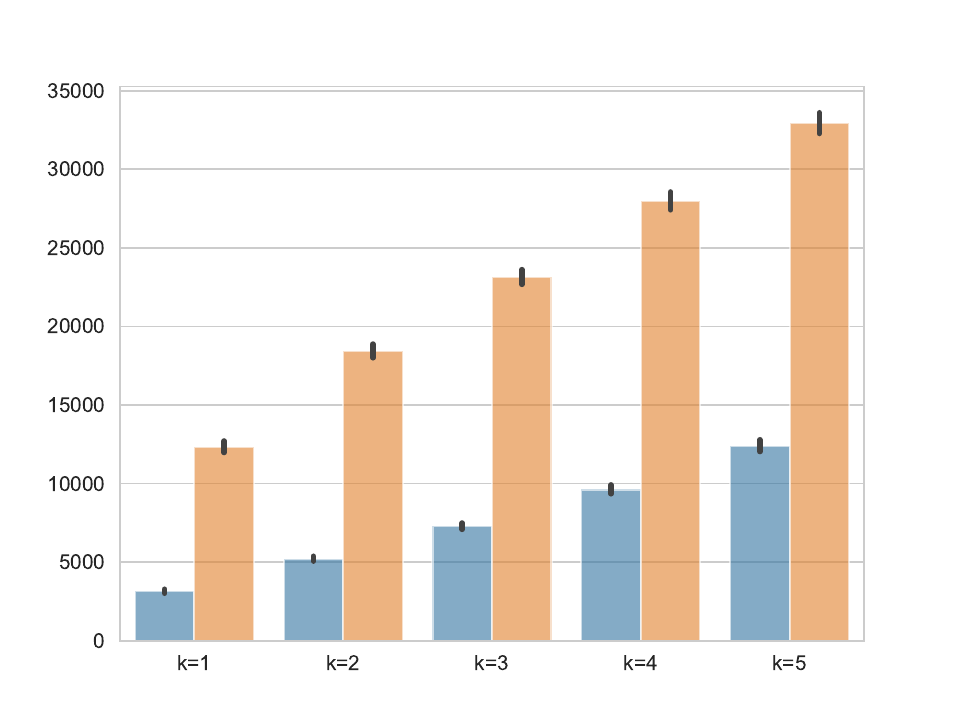}
         \caption{$K=50, D=2, \eps_1=0.1$}
         \label{fig:appx-comp-ape-heuristic-K100k5-eps-01}
     \end{subfigure}
        \caption{Average sample complexity on 2000 random instances (left) and 500 random instances (right).}
        \label{fig:appx-comp-ape-heuristic}
\end{figure}

We set $\delta=0.1$ and we compare both algorithms on 2 type of randomly uniformly generated Bernoulli instances. For the first type we set $K=10, D=2$ and for the second one, we set $K=50, D=2$. For the instances with $K=10$ we set $\eps_1 = 0.05$ and run both algorithms on 2000 random Bernoulli instances with. For the second type of instances $K=50$, we set $\eps_1 = 0.1$ and we benchmark the algorithms on 500 random instances.  The average size of the Pareto set was  2.90 (for $K=10, D=2$) and 4.51 ($K=50, D=2$). 
\autoref{fig:appx-comp-ape-heuristic} shows the average sample complexity of the algorithms for different values of $k \in \{1, \dots, 5\}$. We can observe that the difference between $\eps_1$-APE-$k$ and \algname for the 5 values reported is more important for $K=50$ than for $K=10$. Put together, these experiments show that our algorithm is still preferable for the $k$-relaxation.

\subsubsection{Comparison to some BAI algorithms}
We evaluate the performance of \algname for Best Arm Identification ($D=1$) on two randomly generated instances: one with $K=5$ and means (rounded)
$\cX_1 := (0.25, 0.16, 0.87, 0.22, 0.98)$, and the second one with $K=10$ and means (rounded) $ \cX_2 := (0.43, 0.33, 0.56, 0.85, 0.20,
       0.93, 0.70, 0.82, 0.56, 0.78) $ We use the instantiation of \algname with confidence bonuses on pair of arms {but without the possible improvement invoked in \autoref{rmk:improve_ape_for_BAI}}. 
UGap and LUCB are implemented with the tightest known (to our knowledge) confidence bonus taken from \cite{katz-samuels_top_2019} (in spirit of the finite-time {law of the iterated logarithm}\cite{jamieson_lil_2013}). LUCB++ is used with the improved scheme given in \cite{simchowitz_simulator_2017}.
We set $\delta=0.01$ but the empirical error was way smaller.  The results are averaged over $1000$ independent trials. 

On the \texttt{y}-axis is the sample complexity in units of (an approximation of) the lower bound of BAI for Gaussian rewards with $\sigma=\nicefrac12$\footnote{as Bernoulli distributions are $\nicefrac12$-subgaussian} (\cite{kaufmann_complexity_2014, simchowitz_simulator_2017}) :  
$$ H\log\left(\frac{1}{2.4\delta}\right)\text{ with }  H =\sum_{i=1}^K \frac1{2\Delta_i^2}.$$

\begin{figure}[H]
     \centering
     \begin{subfigure}[b]{0.49\textwidth}
         \centering
    \includegraphics[width=\textwidth]{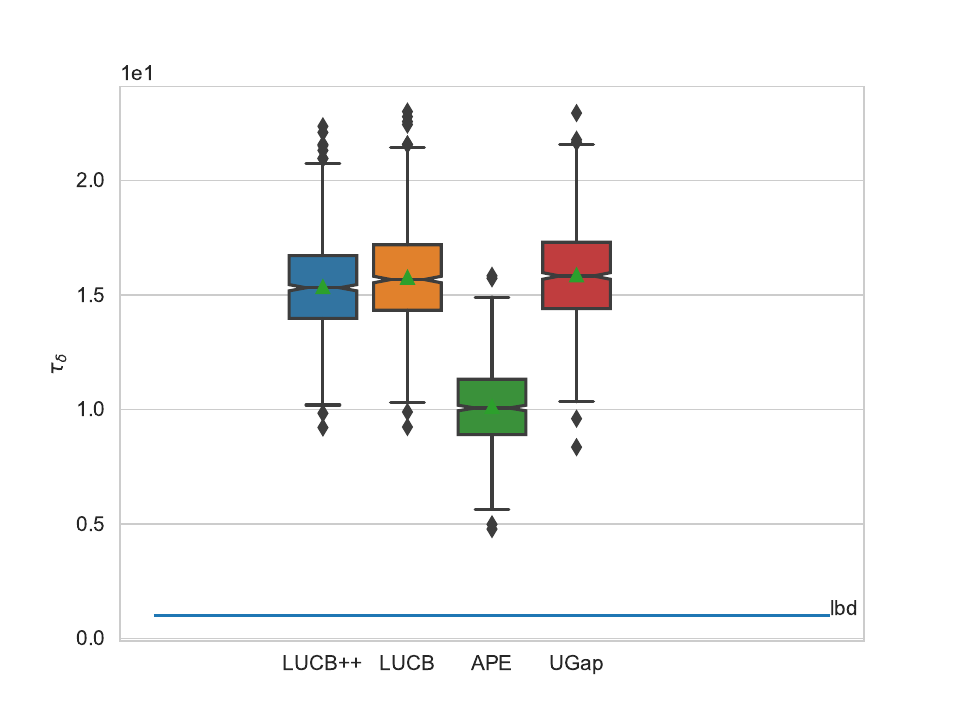}
         \caption{$H = 94.84$}
         \label{fig:appx-F-results-bai-1}
     \end{subfigure}
     \hfill
     \begin{subfigure}[b]{0.49\textwidth}
         \centering
    \includegraphics[width=\textwidth]{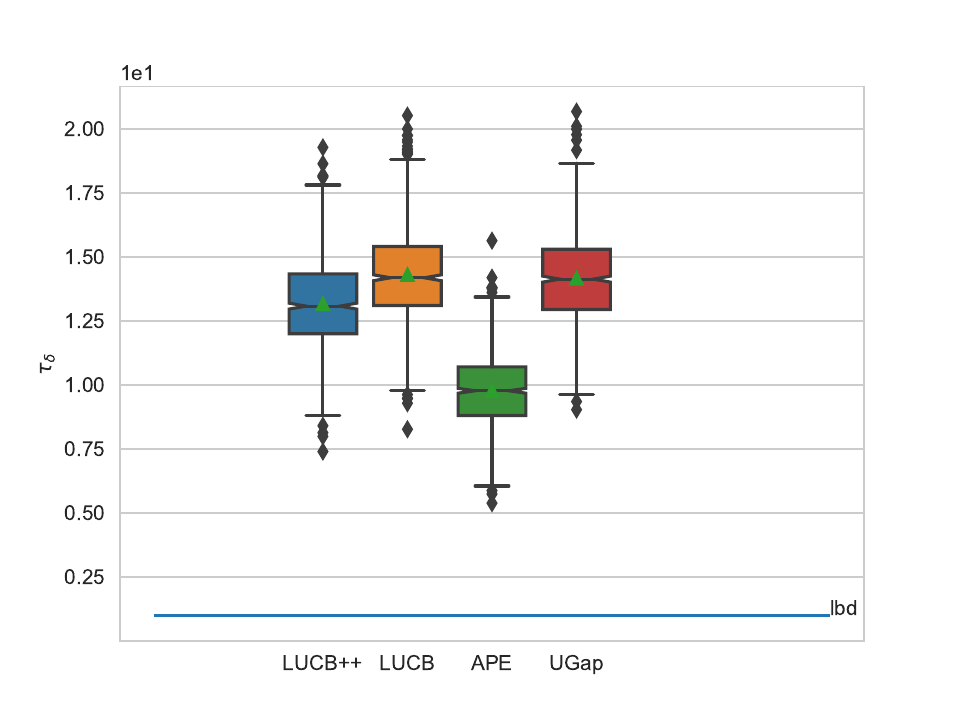}
         \caption{$H = 223.20 $}
         \label{fig:appx-F-results-bai-2}
     \end{subfigure}
        \caption{Empirical sample complexity expressed in units of the lower bound $H\log(1/2.4\delta)$ (blue line) on a random instance with $K=5$ (left)  and $K=10$ (right), $\delta=0.01$. The blue line has a coordinate of 1 on the \texttt{y-}axis (lower-bound).}
        \label{fig:appx-F-results-bai}
\end{figure}
\end{document}